%% file: main.tex
\documentclass[acmlarge]{acmart}

\AtBeginDocument{%
  \providecommand\BibTeX{{%
    \normalfont B\kern-0.5em{\scshape i\kern-0.25em b}\kern-0.8em\TeX}}}

\setcopyright{acmcopyright}
\acmJournal{IMWUT}
\acmYear{2021} 
\acmVolume{5} 
\acmNumber{1} 
\acmArticle{24} 
\acmMonth{3} 
\acmPrice{15.00}
\acmDOI{10.1145/3448125}

\usepackage{hyperref}
\usepackage{url}
\usepackage{graphicx}
\usepackage{enumerate}
\usepackage{ragged2e}
\usepackage{threeparttable} 

\usepackage[caption=false,font=normalsize]{subfig}
\usepackage{xspace}
\usepackage[ruled]{algorithm2e}
\usepackage{multirow, makecell}
\usepackage{enumitem}

\newcommand{\ie}{\emph{i.e.},\xspace}
\newcommand{\vs}{\emph{v.s.}\xspace}
\newcommand{\eg}{\emph{e.g.}\xspace}

\newcommand{\etal}{\emph{et al.}\xspace}

\newcommand\equref[1]{Eq.(\ref{#1})}

\newcommand{\systemname}{{\sf AdaSpring}\xspace}
\newcommand{\systemnameposs}{{\sf AdaSpring's}\xspace}

\newcommand\rev[1]{\textcolor{black}{#1}}

\begin{document}

\title{\justifying{AdaSpring: Context-adaptive and Runtime-evolutionary Deep Model Compression for Mobile Applications}}

\author{Sicong Liu}
\affiliation{%
  \institution{Northwestern Polytechnical University}
  \department{School of Computer Science}
  \city{Xi'an}
  \country{China}
}

\author{Bin Guo}
\affiliation{%
  \institution{Northwestern Polytechnical University}
  \department{School of Computer Science}
  \city{Xi'an}
  \country{China}
  \thanks{Corresponding author: guob@nwpu.edu.cn}
}

\author{Ke Ma}
\affiliation{%
  \institution{Northwestern Polytechnical University}
  \department{School of Computer Science}
  \city{Xi'an}
  \country{China}
}

\author{Zhiwen Yu}
\affiliation{%
  \institution{Northwestern Polytechnical University}
  \department{School of Computer Science}
  \city{Xi'an}
  \country{China}
}

\author{Junzhao Du}
\affiliation{%
  \institution{Xidian University}
  \department{School of Computer Science and Technology}
  \city{Xi'an}
  \country{China}
}

\renewcommand{\shortauthors}{Liu et al.}
\renewcommand{\shorttitle}{AdaSpring: Context-adaptive and Runtime-evolutionary Deep Model Compression for Mobile Applications}

\begin{abstract}

There are many deep learning (\eg DNN) powered mobile and wearable applications today continuously and unobtrusively sensing the ambient surroundings to enhance all aspects of human lives. To enable robust and private mobile sensing, DNN tends to be deployed locally on the resource-constrained mobile devices via model compression. The current practice either hand-crafted DNN compression techniques, \ie for optimizing DNN-relative performance (\eg parameter size), or on-demand DNN compression methods, \ie for optimizing hardware-dependent metrics (\eg latency), cannot be locally online because they require offline retraining to ensure accuracy. Also, none of them have correlated their efforts with runtime adaptive compression to consider the dynamic nature of deployment context of mobile applications. To address those challenges, we present AdaSpring, a context-adaptive and self-evolutionary DNN compression framework. It enables the runtime adaptive DNN compression locally online. Specifically, it presents the ensemble training of a retraining-free and self-evolutionary network to integrate multiple alternative DNN compression configurations (\ie compressed architectures and weights). It then introduces the runtime search strategy to quickly search for the most suitable compression configurations and evolve the corresponding weights. With evaluation on five tasks across three platforms and a real-world case study, experiment outcomes show that AdaSpring obtains up to $3.1\times$ latency reduction, $4.2\times$ energy efficiency improvement in DNNs, compared to hand-crafted compression techniques, while only incurring $\leq 6.2ms$ runtime-evolution latency.
%

\end{abstract}

\ccsdesc[500]{Human-centered computing~Ubiquitous and mobile computing systems and tools}

\maketitle

\input{body/intro}
\input{body/related}
\input{body/overview}
\input{body/offline}

\input{body/online}
\input{body/experiment}
\input{body/conclusion}

\begin{acks}
This work was partially supported by the National Key R\&D Program of China (2019YFB1703901), National Science Fund for Distinguished Young Scholars (62025205, 61725205),  National Natural Science Foundation of China (No. 62032020, 61960206008, 62032017), and the Fundamental Research Funds for the Central Universities (No. 3102020QD1005). The authors also thank the anonymous reviewers for their constructive feedback that has made the work stronger.
\end{acks}

\bibliography{acmart}
\bibliographystyle{ACM-Reference-Format}

\end{document}

%% file: body/intro.tex
\section{introduction}
\label{sec:introduction}

%
In recent years, a lot of \rev{ubiquitous devices (\eg smartphones, wearables, and embedded facilities) are integrated with continuously running applications to facilitate all aspects of human lives.}
For example, the smartphone-based speech assistant (\eg ProxiTalk~\cite{bib:yang2019proxitalk}), 
and wearable sensor-enabled activity recognition   (IMUTube~\cite{bib:kwon2020imutube}, MITIER~\cite{bib:chen2020metier}).
Notably, there is a growing trend to bring deep learning (\eg DNN) powered intelligence into mobile devices, which benefits effective data analysis.
%
Besides, due to the increasing user concerns on transmission cost and privacy issues, executing DNN on local devices tends to be a promising paradigm for robust mobile sensing ~\cite{bib:wang2020convergence, bib:lane2015deepear}.
However, it is non-trivial to deploy the computational-intensive DNN on mobile platforms with tightly limited resources (\ie storage, battery).

\begin{figure*}[t]
  \centering
   \subfloat[Compression]{
   \includegraphics[height=0.13\textwidth]{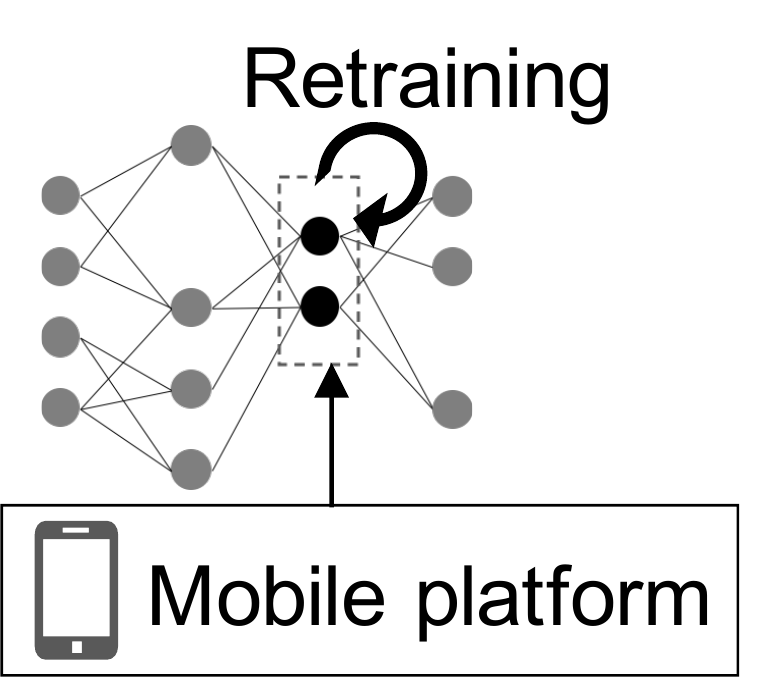}}
   \hfill
   \subfloat[On-demand compression]{
   \includegraphics[height=0.13\textwidth]{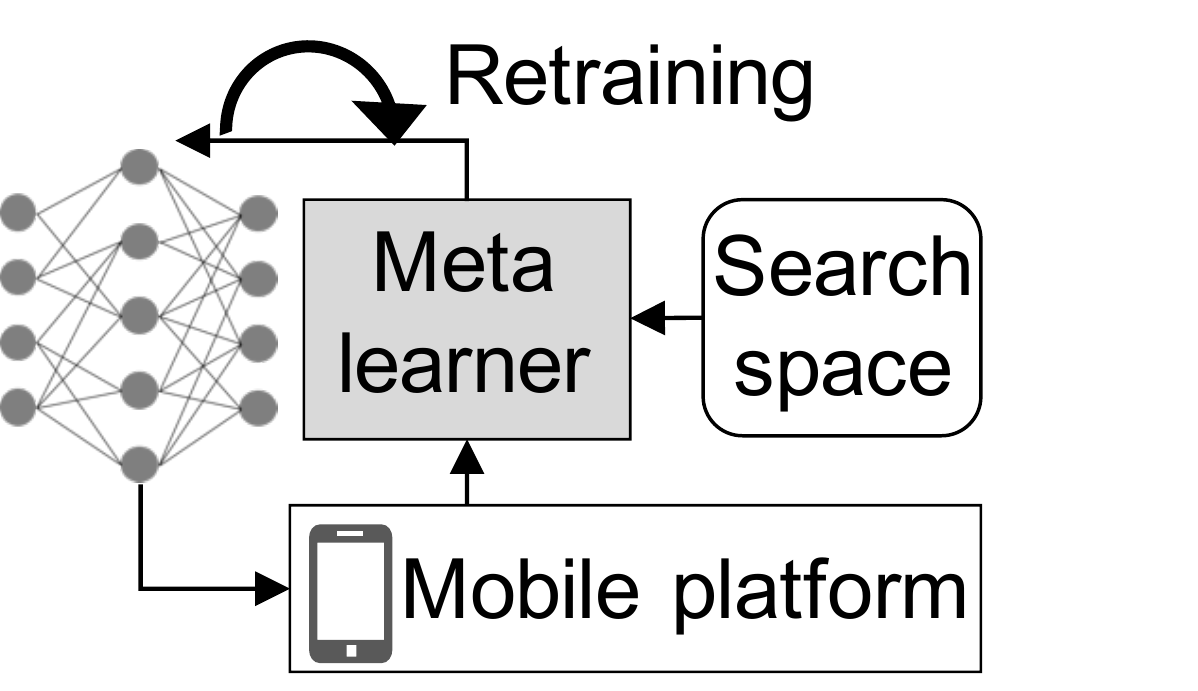}}
   \hfill
   \subfloat[\rev{One-shot NAS}]{
   \includegraphics[height=0.13\textwidth]{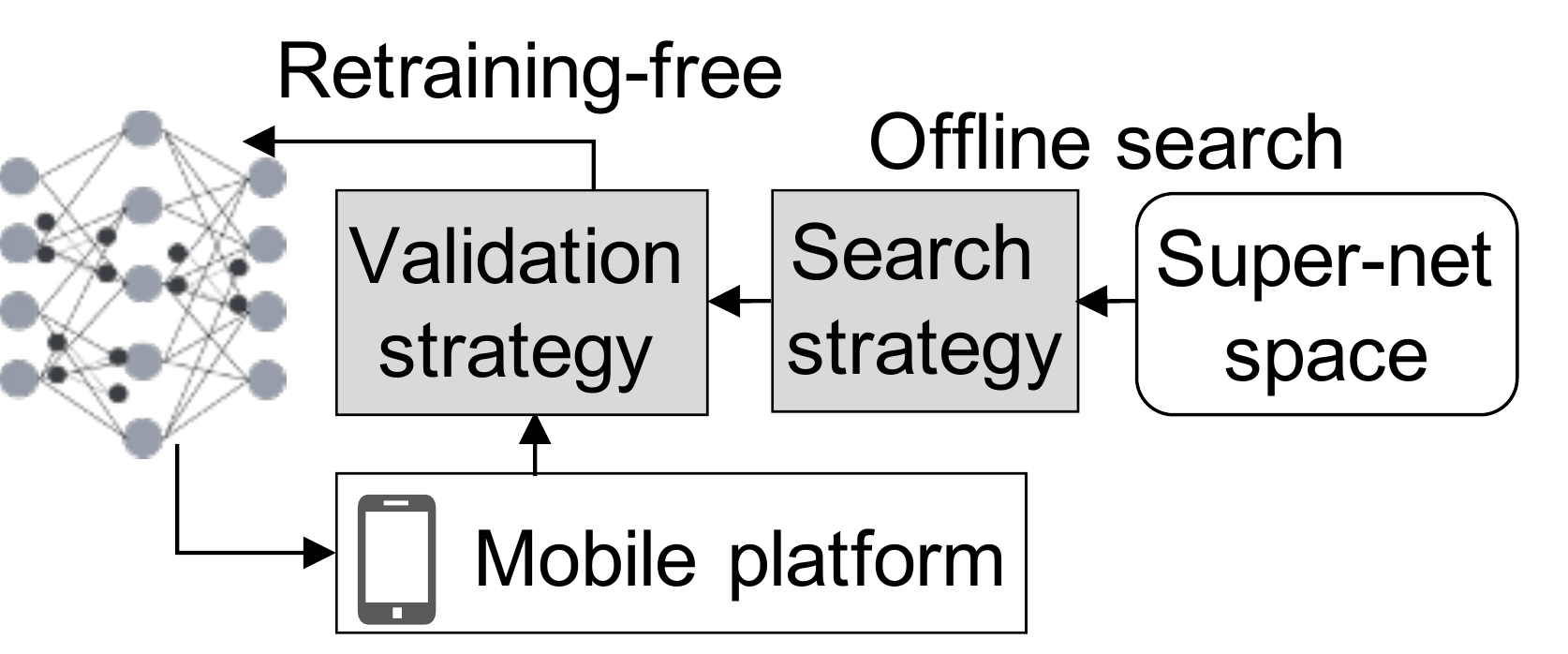}}
  \hfill
  \subfloat[\rev{Runtime adaptive compression}.]{
  \includegraphics[height=0.13\textwidth]{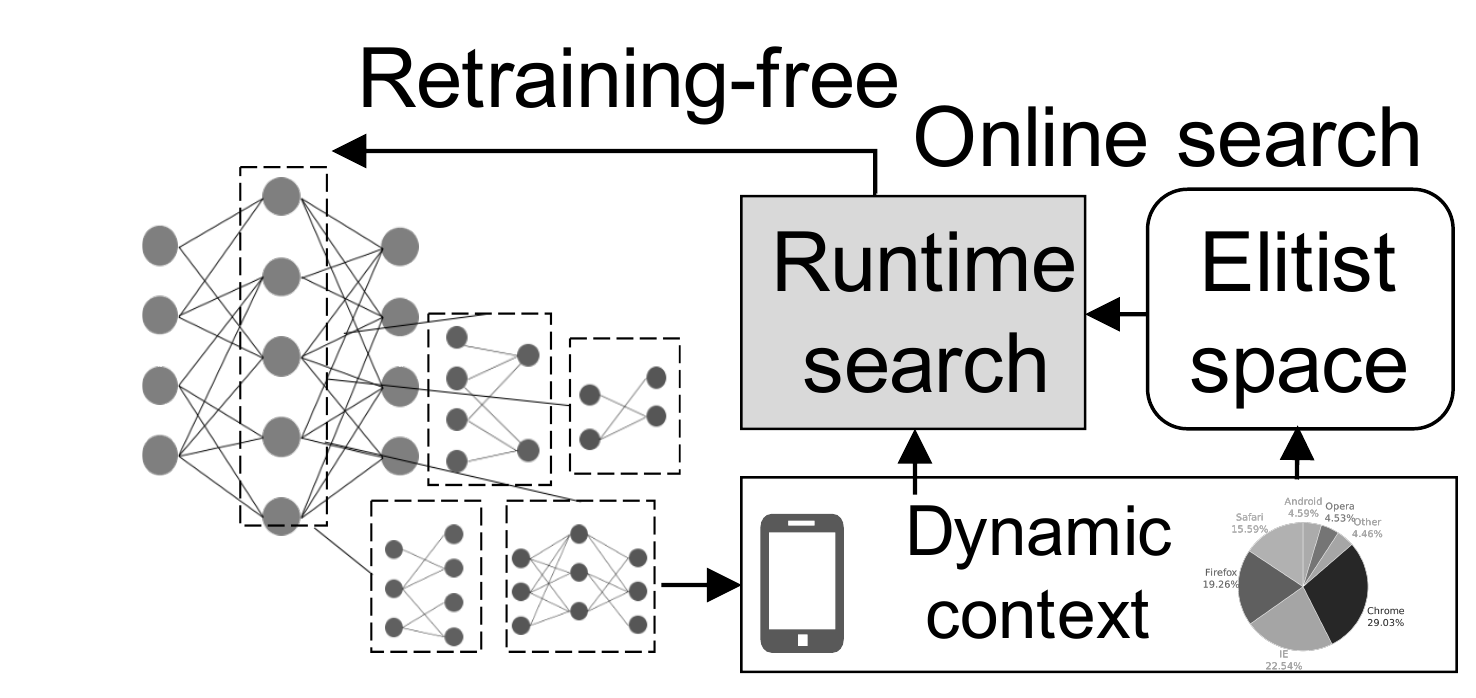}}
\caption{Illustration of various DNN specialization methods: (a)hand-crafted compression relays on manual design, (b)on-demand compression and (c)\rev{one-shot NAS always bring high overhead for searching large-scale candidates}, while (d)\rev{runtime adaptive compression schemes should be locally online (\ie in millisecond level), self-adaptive, and retraining-free}.
}
\label{fig_motivation}
\end{figure*}

Given those challenges, prior works have investigated different DNN specialization schemes to explore the desired tradeoff between application performance (\ie accuracy, latency) and resource constraints (\ie battery and storage budgets).
\textit{Firstly}, as illustrated in Figure \ref{fig_motivation}(a), the \textit{hand-crafted DNN compression} methods, \eg weight pruning~\cite{bib:luo2020autopruner} relay on manual design to reduce the model complexity.
%
They may not suffice to meet diverse performance requirements.
\textit{Secondly}, the \textit{on-demand DNN compression} schemes (see Figure \ref{fig_motivation}(b)), \eg DeepX~\cite{bib:lane2016deepx}, AMC~\cite{bib:he2018amc}, and AdaDeep~\cite{bib:liu2020adadeep}, adopt a trainable meta-learner to automatically find the most suitable DNN compression strategies for various platforms.
They need \textit{offline retraining} to ensure accuracy and update the meta-learner. The  \rev{extra overhead and latency for offline retraining is intolerable for responsive  applications.} 
\textit{Thirdly}, the \textit{one-shot neural architecture search (NAS)} methods (see Figure \ref{fig_motivation}(c)) \rev{pre-train a super-net and automatically search for the best DNN architecture for target platforms \cite{bib:fang2020fast, bib:zoph2018learning,bib:saikia2019autodispnet,bib:cai2019once}.
However, they also render high overhead for \textit{scanning and searching a large-scale candidate space}}. \rev{None of them can work locally online.}

\begin{figure*}[tb]
  \centering
  \includegraphics[width=.98\textwidth]{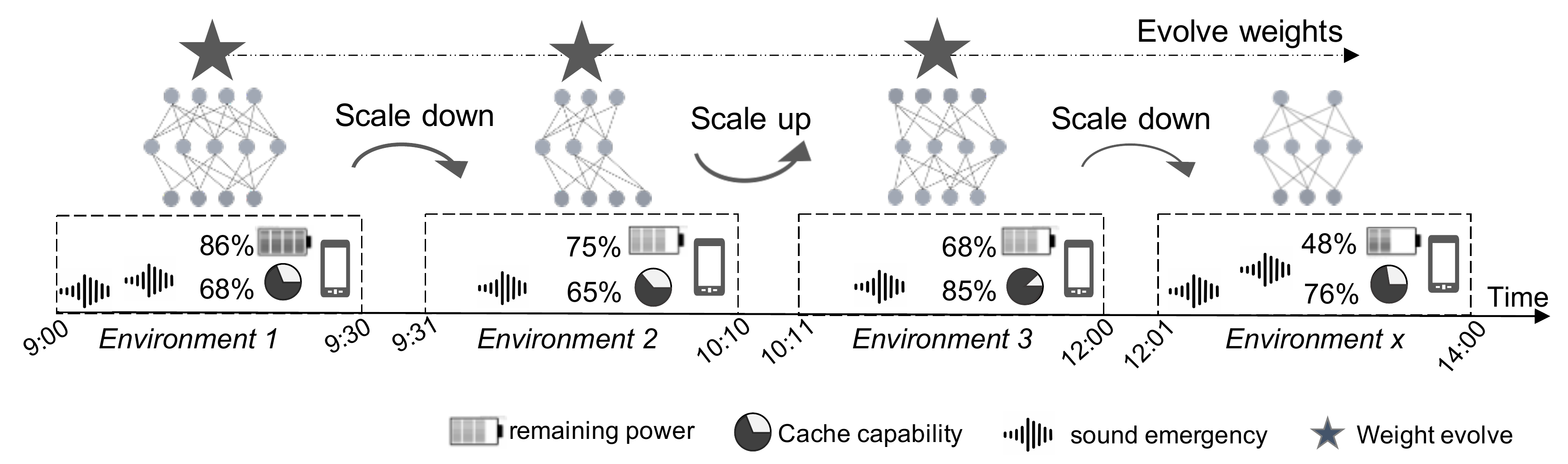}
  \caption{An example of runtime adaptive DNN compression, \ie continually scaling down/up architecture and evolving weights, which is desired by continuously running mobile applications. The dynamic deployment context contains the unpredictable remaining power and memory capability of mobile platforms, and the sound happening frequency in ambient environments.}
  \label{fig_evolution}
\end{figure*}

Despite major advances of existing DNN compression techniques, none of them have correlated their efforts with \textit{runtime adaptive compression}, to consider the \textit{dynamics} of the deployment context in continuously running applications.
We have identified that a self-adaptive, retraining-free, and fast framework for runtime adaptive DNN compression (see in Figure\ref{fig_evolution}(d)) is necessary yet challenging.
As we have illustrated in Figure \ref{fig_motivation}(d), the DNN deployment context often exhibits high dynamics and unpredictability in practice. 
And the dynamic changes in the deployment context will further lead to varying performance demands on DNN compression.
Specifically, we identify the dynamic context to mainly include the time-varying hardware capabilities (\eg storage, battery, processor), the DNN active execution time, the agnostic inference frequency triggered by real environments, and the unpredictable resource contention imposed by other Apps.

Figure~\ref{fig_evolution} shows an example in which a user carries a smartphone-based hearing assistant App (\eg UbiEar~\cite{bib:sicong2017ubiear}) to sense the ambient acoustic event of interest continuously. 
During its use, the smartphone’s battery is dynamically consumed by the DNN execution, the memory access, the microphone sampling, and the screen with unpredictable frequency, which further characterize the dynamic energy constraints for the deployed DNN.
And the storage unit (\eg L2-Cache) is also dynamically occupied by other applications, resulting in various storage budgets for DNN parameters.
Both the mobile developer and user face a problem: \textit{how to automatically and effectively re-compress DNN at runtime to meet dynamic demands?} And they face the following two challenges:
\begin{itemize}
\item \textit{Firstly}, it is non-trivial to \textit{continually scale up/down} the DNN compression configurations, including both architectures and weights, to meet the dynamic optimization objectives on multiple DNN performance (\ie accuracy, latency, energy consumption) on-the-fly. 
%
%
This is because most DNN compression methods are \textit{irreversible} to scale up DNN again, \ie recover fine details of the DNN architecture and weight, from a compressed/pruned model. 
And the weight evolution is always limited by \textit{offline retraining}.
%
\item \textit{Secondly,} it is intractable to provide an efficient and effective solution to the \textit{runtime optimization} problem.
To tailor this problem, \textit{quickly searching} for the suitable compression techniques \rev{from an \textit{elite candidate set}} and \textit{efficiently evolving} weights without model retraining are required.
Moreover, it is difficult to systematically balance the compromising of multiple conflicting and interdependent performance metrics (\eg latency, storage and energy efficiency) by merely tuning the DNN compression methods.

\end{itemize}

In view of those challenges and limitations, we present \systemname, a context-adaptive and runtime-evolutionary deep model compression framework. 
It continually controls the compromising of multiple performance metrics by re-selecting the proper DNN compression techniques.
To formulate the dynamic context, we formulate the runtime tuning of compression techniques by a dynamic optimization problem (see \equref{equ_opt} in $\S$ \ref{sec:problem}).
In that, we model the dynamic context by a set of time-varying constraints (\ie accuracy loss threshold, latency and storage budgets, and the relative importance of objectives).
And then, we present a heuristic solution.
%
In particular, to eliminate the runtime retraining cost, we decouple offline training from the online adaptation by putting weight tuning ahead in the training of a self-evolutionary network (see $\S$ \ref{sec:offline}).
Furthermore, we present an efficient and effective search strategy. \rev{It involves an elite and flexible search space (see $\S$ \ref{subsec:space})}, the progressive shortest candidate encoding, and the Runtime3C search algorithm (see $\S$ \ref{subsec:search}) to boost the locally online search efficiency and quality.
The main contributions of this work are summarized as follows.
\begin{itemize}
    \item To the best of our knowledge, \systemname is the first context-adaptive, and self-adaptive DNN compression framework to \textit{continually} shrink model architectures and evolve the corresponding weights by \textit{automatically} applying the proper retraining-free compression techniques on-the-fly.
    And it trains a self-evolutionary network to \rev{ synergize the multi-scale compression operators' weight recycle and decouple offline training from online compression.}

    \item \systemname presents an efficient runtime search strategy to optimize the runtime adaptive compression problem. 
    It introduces the \rev{elite and flexibly combined compression operator space, the fast \textit{Runtime3C search} algorithm, and a set of speedup mechanisms to boost the search efficiency and quality,} at runtime, while avoids explosive combination.
    %
    %
    
    \item Using five mobile applications 
    across three platforms and a real-world case study of DNN-powered sound recognition on NVIDIA Jetbot, 
    extensive experiments showed the advantage of \systemname to continually optimize DNN configurations. It adaptively adjust the compression configurations to tune energy cost by $1.6mJ\sim5.6mJ$, latency by $1.3ms \sim 10.2ms$, and storage by $201KB \sim 1.9MB$, with $\leq 2.1\%$ accuracy loss. And the online evolution latency of compression configurations to meet dynamic contexts is $\leq 6.2ms$. 
\end{itemize}

In the rest of the paper, we review the related work in $\S$ \ref{sec:related}, present the system overview in $\S$ \ref{sec:problem}, and elaborate the \systemname design in $\S$ \ref{sec:online} and $\S$ \ref{sec:offline}.
We report an evaluation of \systemname in $\S$ \ref{sec:experiment} and conclude in $\S$ \ref{sec:conclude}. 

%% file: body/related.tex
\section{Related Work}
\label{sec:related}
Our work is inspired by and closely related to the following works. 


\textbf{DNN Compression for Ubiquitous Mobile Applications}.
There is a promising trend to bring deep learning powered intelligence to mobile and embedded devices (\eg smartphones, wearables, IoT) for enhancing all aspects of human lives.
%
%
%
%
%
such as smartphone-based speech assistant (\eg ProxiTalk~\cite{bib:yang2019proxitalk}), smarthome device-enabled sound detection~\cite{bib:bhattacharya2020countering}, and wearable-based activity recognition (\eg MITIER~\cite{bib:chen2020metier}).
%
%
%
And recent research has demonstrated the potential of feeding DNNs into resource-constrained mobiles~\cite{bib:cheng2017survey} by using DNN compression techniques, including parameter pruning~\cite{bib:he2017channel}, 
 sharing~\cite{bib:wu2018deep}, and quantification~\cite{bib:zhu2018bayesian}, compact component~\cite{bib:iandola2016squeezenet, bib:howard2017mobilenets}, and model distillation~\cite{bib:chen2017learning}, 
%
%
%
%
However, we note that few efficient compression techniques are dedicated to optimizing the application-driven system performance (\eg energy efficiency). 
For example, the recent work~\cite{bib:jha2019ramifications} argues that the platform-aware SqueezeNet~\cite{bib:iandola2016squeezenet} and SqueezeNext~\cite{bib:gholami2018squeezenext} merely reduce parameter size or MAC amount which do not necessarily lead to reduced energy cost or latency~\cite{bib:yang2017designing}. 
%
%
Also, all of these techniques need several epochs of model retraining to ensure accuracy, thereby they can not be locally online.
%
Instead, \systemname decouples DNN training from online adaptive compression, and consider both the dynamic arithmetic intensity of parameters and activations for guiding the best specialization of convolutional compression configurations.


%
%

\textbf{On-demand DNN Computation for Diverse User Demands}.
There have been two categories of on-demand DNN computation adjustment methods: on-demand DNN compression~\cite{bib:liu2018demand} \cite{bib:luo2020autopruner} 
and on-demand DNN segmentation~\cite{bib:zhao2018deepthings}.
They specialize DNN computation offline to meet diverse hardware resource budgets (\eg battery, memory, and computation) and application demands (\eg input diversity).
To satisfy resource budgets, 
He \etal \cite{bib:he2018amc} adopt reinforcement learning to adaptively sample the design space.
%
Shuochao \etal \cite{bib:yao2017deepiot} use a recurrent model to control the adaptive compression ratio of each layer.
%
%
Singh \etal ~\cite{bib:singh2019play} introduce a min-max game to achieve maximum pruning with minimal accuracy drop.
%
These methods, however, need extra offline training to update the meta-controller for on-demand adjustment.
%
%
%
%
Zhao \etal~\cite{bib:zhao2018deepthings} present the adaptively distributed execution of CNN-based  applications on resource-constrained IoT edge clusters.
However, because of mobile platforms' mobility and opportunistic connectivity, distributed DNN inference in mobile clusters is not robust yet.
Built upon these efforts, \systemname is the first to enable runtime and adaptive DNN evolution locally without requiring Wi-Fi/cellular networks to connect with other platforms while achieve competitive performance.


\textbf{Dynamic Adaptation of DNN Execution}.
Prior works have investigated the run-time adaptation of DNN execution to adapt to diverse inputs from two directions: dynamic selection of inference path~\cite{bib:wu2018blockdrop} or network variant~\cite{bib:teerapittayanon2016branchynet}.
%
Wu \etal~\cite{bib:wu2018blockdrop} adaptively choose which residual blocks to execute during inference to  reduce computation without degrading  accuracy.
Teerapittayanon \etal~\cite{bib:teerapittayanon2016branchynet}
propose a multi-branch network to allow inference adaptively exit from early branches.
%
%
%
%
Han \etal \cite{bib:han2016mcdnn} adaptively select model variants to optimize accuracy and satisfy resource constraints (\eg memory and energy).
Gao \etal \cite{bib:gao2018dynamic} propose feature boosting and suppression method to predictively amplify salient convolutional channels and skip unimportant ones.
However, these methods highly depend on the pre-defined design space of alternative execution paths and variants, but it is prohibitive to specify all of them before deploying models into agnostic mobile contexts.
\systemname dynamically select and combine the proper compression operators to flexibly shrink the model configurations from multiple scaling dimensions at runtime.

\textbf{Fast and Platform-aware Neural Architecture Search}.
Recent studies have verified the potential of leveraging neural architecture search (NAS) framework to automate neural architecture specialization for mobile applications, from two aspects.
Firstly, the Fast-NAS aims to automatically specialize neural architecture for different performance demands (\eg computation amount, parameter size), using as little search cost as possible~\cite{bib:ren2020comprehensive}.
%
To speedup the search, researchers have investigated the modular search strategy~\cite{bib:cai2018proxylessnas,bib:zhong2018practical,bib:liu2017hierarchical}, differentiable search strategy~\cite{bib:liu2018darts, bib:jiang2019improved, bib:chen2019progressive, bib:zhou2020theory}, and super-network  search strategy~\cite{bib:fang2020fast, bib:bender2018understanding, bib:cai2019once}.
%
For example, 
%
Liu \etal \cite{bib:liu2018darts} relax the search space to be continuous, so that it can be optimized by gradient descent, using orders of magnitude less search cost. 
%
%
Cai \etal \cite{bib:cai2019once} trains an once-for-all (OFA)  super-network \rev{that supports $2\times {10^{19}}$ diverse variant-network search.}
\rev{We note that the OFA super-network includes some redundant and invalid variant-networks, which is not elite and incurs a high search cost.} (As we will discuss in $\S$ \ref{sec:online}.)
Secondly, unlike general NAS that only optimize for model-relative metrics, such as FLOPS, the platform-aware NAS alao incorporates 
platform-relative metrics (\eg latency) into optimization objectives.
Such as Mingxing \etal ~\cite{bib:tan2019mnasnet} explicitly incorporate latency into the NAS objective to identify a mobile CNN model.
%
Xiaoliang \etal \cite{bib:dai2019chamnet} propose an efficient search algorithm aided by efficient accuracy and hardware resource predictors.
However, above methods still sacrifice high overhead to obtain the ranking of candidate architectures based on their performance on validation sets. 
And they donot accurately consider the energy consumption improvement target since the energy efficiency measurement is not straightforward on different platforms with dynamic nature.
%
%
%
Depart from existing efforts, \systemname treats the retraining-free compression operator (illustrated in $\S$ \ref{subsec:space}) as a new ensemble to be tuned by automated macro-NAS. 
It trains a self-evolutionary network(see $\S$ \ref{sec:offline}) at design time to decouple model retraining and adaptive compression.
Besides, it present a set of mechanisms to boost search efficiency (\ie at millisecond level) and quality during dynamic inference.
Notably, \systemname leverages the dynamically measured hardware-relative metrics (\ie arithmetic intensity of parameter and activation) to derive a guiding selection, which also prevent the explosive combination (see $\S$ \ref{sec:online}).

%% file: body/overview.tex
\section{Overview}
\label{sec:problem}
This section starts with problem analysis and then presents an overview of \systemname design.

\begin{figure*}[t]
  \begin{minipage}{0.48\linewidth}
  \centering
  \includegraphics[height=0.99\textwidth]{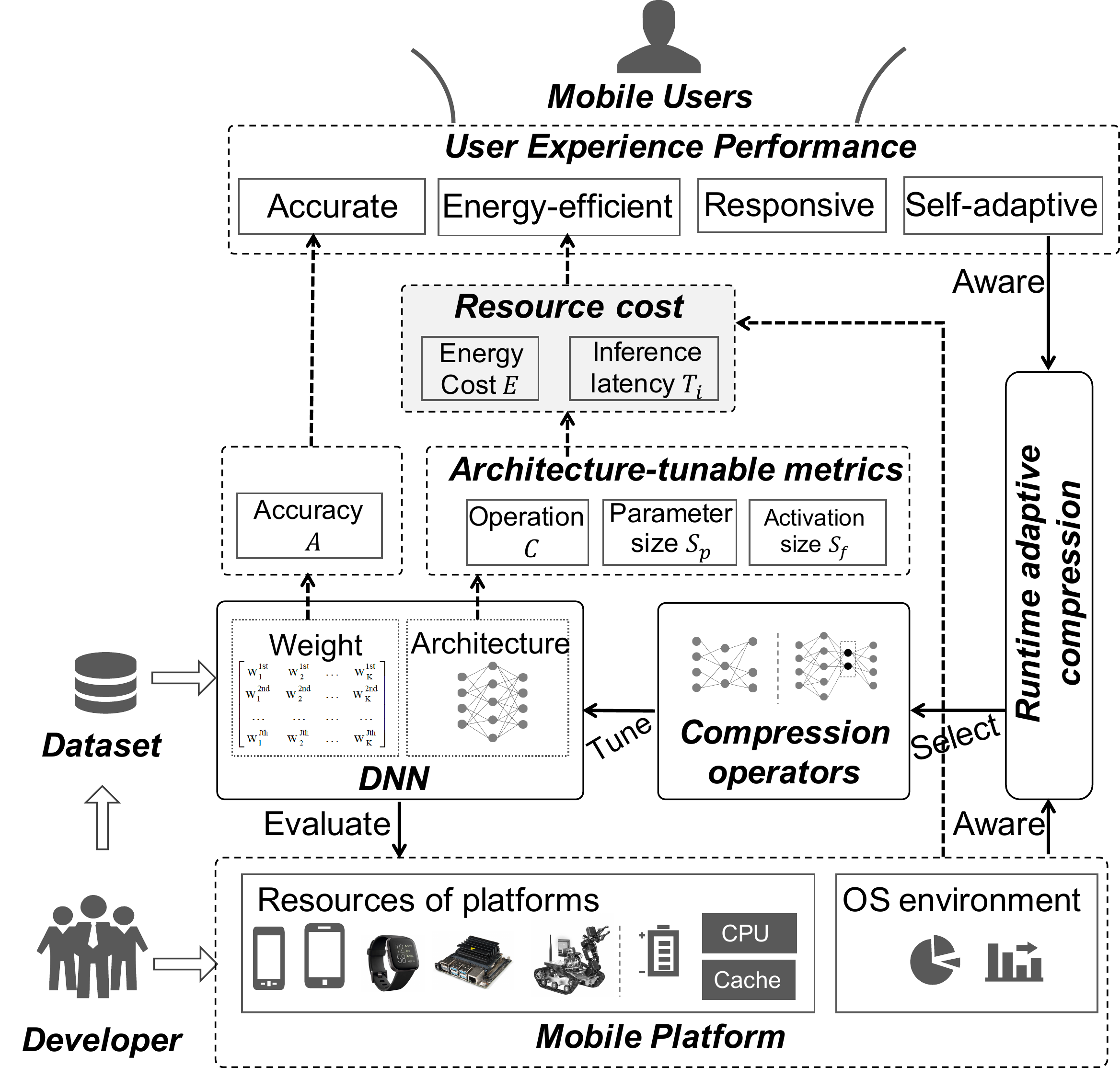}
  \caption{\systemname aims to automate the compression operator selection to systematically tune the user experience.}
  \label{fig_metric}
  \end{minipage}
   \hfill
  \begin{minipage}{0.48\linewidth}
  \centering
\includegraphics[height=.99\textwidth]{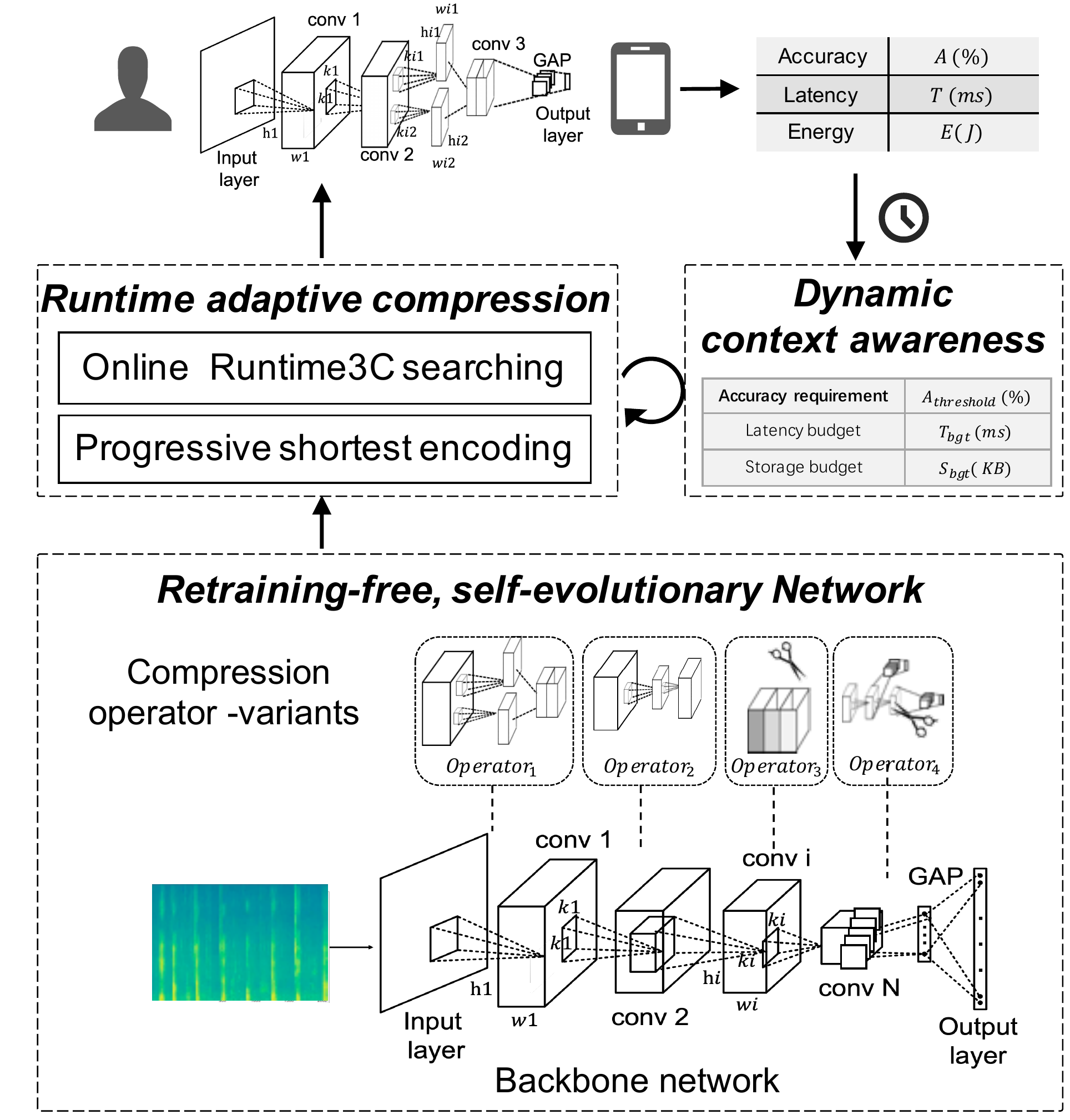}
  \caption{\systemname continually optimizes DNN performance metrics by selecting proper compression operators.
  }
  \label{fig_overview}
  \end{minipage}
\end{figure*} 

\subsection{Problem Study}

Due to the dynamic natures of DNN deployment context, we aim to continually tune the DNN compression configurations to directly/indirectly optimize the application-driven system performance (\ie accuracy, energy efficiency, latency).
The hybrid dependency of multiple platform-relative performance metrics and DNN-dependent metrics are shown in Figure \ref{fig_metric}.
%
%
To further understand the performance requirements of DNN for the continuously running mobile applications, we ask $60$ mobile users and  $10$ Android developers to rate the importance of different DNN performance aspects on mobiles.
And we summarize the results as our design goals.
Specifically, a DNN for continuously running mobile Apps needs to fulfill the following requirements:
\begin{itemize}
    \item \textbf{Accurate}: the DNN is accurate enough to guarantee a high-quality task. The model weights at different scales are well-trained to represent the generic information of recognition objects.
    
    \item \textbf{Responsive}: the complexity of the DNN should be controllable to satisfy diverse user demands on latency constraints, especially on low-end (\eg CPU-powered) mobiles. 
    
    \item \textbf{Energy-Efficient}: the energy consumption of the DNN should be continually optimized, which is the bottleneck metric for continuously sensing applications~\cite{bib:yang2017designing}.
    
   \item \textbf{Runtime-evolutionary}: both the DNN architecture and parameter weights are runtime-evolutionary to meet the dynamic deployment context for continually optimizing the above three requirements (\ie accurate, responsive, energy-efficient) at runtime. 
   %
   %

\end{itemize}
Unfortunately, none of previous efforts satisfy all these requirements (as discussed in $\S$ \ref{sec:related}).
To this end, this paper proposes \systemname, a context-adaptive and runtime-evolutionary DNN compression framework to automatically optimize the requirements mentioned above, which are closely related to the user experience.

\subsection{Optimization Formulation}
As shown in Figure \ref{fig_metric}, \systemname intends to provide a systematic method to automatically select the compression operator combination for tuning the above conflicting and interdependent performance metrics.
Mathematically, \systemname explores an efficient solution to the following dynamic optimization problem:
\begin{eqnarray}\label{equ_opt}
\mathop{arg min} \limits_{\delta_i \in \mathrm{ \Delta }} &  \lambda_1(t) Norm(A(\Omega)- A(\Omega( \delta_i)))  - \lambda_2(t) Norm( E(\Omega(\delta_i)) \nonumber \\
\text{s.t.} & A(\Omega)- A(\Omega( \delta_i)) \leq A_{threshold}(t) ,  \,\,  T(\Omega(\delta_i)) \leq T_{bgt}(t) , \,\, S(\Omega(\delta_i)) \leq S_{bgt}(t)
\label{equ:opt}
\end{eqnarray}
where $\mathrm{ \Delta }$ represents the set of all optional convolutional compression operators (as enumerated in $\S$ \ref{subsec:space}).
Given an backbone-net architecture $\Omega$, $\Omega(\delta_i)$ represents the re-configured model architecture compressed by the selected compression operator $\delta_i$.
$A$, $E$, $T$ and $S$ denote the measured accuracy, energy efficiency, latency, and memory footprint of a given model running on the target mobile platform.
The two objectives on $A$ and $E$ are combined by relative importance coefficients $\lambda_1$ and $\lambda_2$, which dynamically depend on the platform's remaining battery.
We express the dynamic deployment contexts as a set of time-varying constraints, \ie the threshold of accuracy loss $A_{threshold}(t)$, the latency budget $T_{bgt}(t)$, the storage budget $S_{bgt}(t)$, and relative importance coefficients of objectives ($\lambda_1(t)$, $\lambda_2(t)$).
%
%
The latency budget $T_{bgt}(t)$ is application-specified. And the storage budget $S_{bgt}$(t) is platform-imposed.
For example, reducing the model size $S$ to satisfy the budget of L2-Cache $S_{bgt}$ helps to fit it into the on-chip memory and avoids the expensive off-chip access.
We note that $Norm(.)$ is a normalization operation for  objective aggregation, \eg $log(.)$.
We then propose a heuristic optimization solution as adjusting the model architecture for satisfying dynamic performance requirements.
%
In particular, the model architecture $\Omega(\delta_i)$ can directly determine both $S$ and $A$ (see Figure \ref{fig_metric}).
While the quantification of hardware-dependent metrics $E$ and $T$ are not straightforward.
Therefore, \systemnameposs goal turns to adaptively select the compression operator combination $\delta_i$ from a discrete set of all possible combinations $\mathrm{ \Delta }$, so that it can directly/indirectly tune model performance metrics.

\subsection{\systemname Framework}
The above challenging problem motivates the \systemname design.
%
As shown in Figure~\ref{fig_overview}, the \systemname framework consists of a \textit{self-evolutionary network}, a \textit{runtime adaptive compression} block, and a \textit{dynamic context awareness} block. 
\textit{(i)} The \textit{self-evolutionary network} is an ensemble of a backbone-net and multiple retraining-free compression operator-variants, which enables weight recycle between numerous variants while avoiding catastrophic interference.
We initialize the backbone-net's hyperparameters at design time using an on-demand DNN generation framework, \ie AdaDeep~\cite{bib:liu2020adadeep}, for satisfying mobile application performance demands on a target platform.
%
\textit{(ii)} The \textit{runtime adaptive compression} block is capable of selecting a deterministic optimal combination of compression operators for reconfiguring and evolving the backbone-net at runtime. 
And \textit{(iii)} the \textit{dynamic deployment context awareness} block detects the evolution demands and triggers the \textit{runtime adaptive compression} block.
The triggering station can be modeled as the noticeable context changes or by a pre-defined frequency (\eg time slice) for continuously running Apps in regular days.

%% file: body/offline.tex
\section{Retraining-free and Self-evolutionary Network Design}
\label{sec:offline}
This section presents the design of the retraining-free and self-evolutionary network.
The self-evolutionary network consists of a high-performance backbone network and multiple compression operator-variants.
%

%

\begin{figure}[t]
  \centering
  \subfloat[Operator $\delta_1$.]{
  \includegraphics[height=0.1\textwidth]{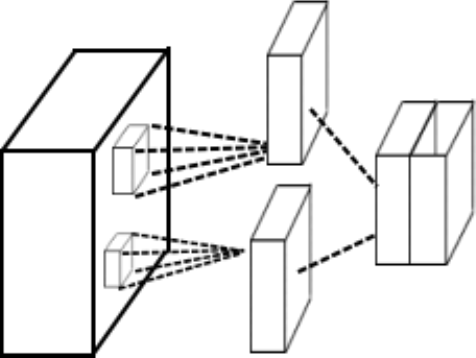}}
  \hspace{9mm}
   \subfloat[Operator $\delta_2$.]{
   \includegraphics[height=0.1\textwidth]{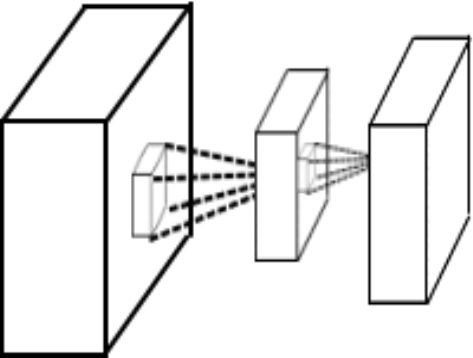}}
\hspace{9mm}
   \subfloat[Operator $\delta_3$.]{
   \includegraphics[height=0.1\textwidth]{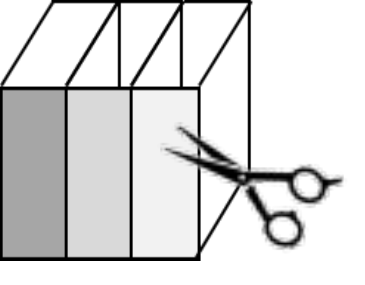}}
\hspace{9mm}
   \subfloat[Operator $\delta_4$.]{
   \includegraphics[height=0.1\textwidth]{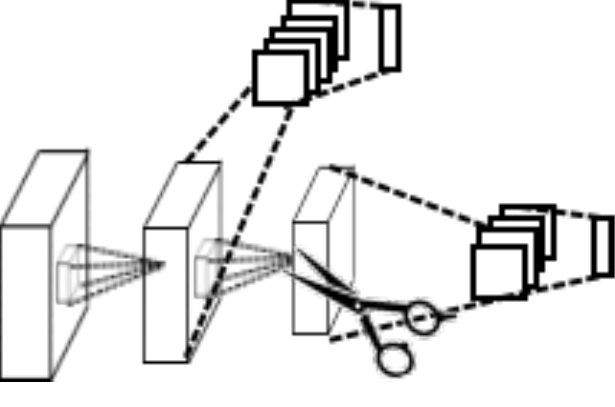}}
\caption{Four categories of compression operators that synthesize multiple scaling dimensions.}
\label{fig_operator}
\end{figure}

%


\subsection{Compression Operators}
\label{subsec:operator}
This paper focus on the configuration optimization of convolutional architecture, operations, and activations.
Because recent successful DNN models tend to shift more parameters on convolutional layers and use fewer fully-connected layers~\cite{bib:wu2018deep, bib:chen2016compressing, bib:cai2019once}.
Built upon the existing compression experience, we propose the following alternative convolutional compression operators that synthesize multiple scaling dimensions (\eg width, depth, and connection).
\begin{itemize}
    \item Compression operator $\delta_1$: \textit{multi-branch channel merging}  techniques(\eg Fire block~\cite{bib:iandola2016squeezenet}) increase the model depth with less parameters by
    replacing one conv layer using two conv layers (\ie squeeze layer and expand layer) which is elaborately designed to decrease the kernel size and channel size per unit.
    \item Compression operator $\delta_2$: \textit{low-rank convolution factorization techniques} (\eg SVD-based~\cite{bib:wu2018deep}, sparse coding-based~\cite{bib:bhattacharya2016sparsification} factorization, or depth/group-wise convolution~\cite{bib:li2019dabnet}) decompose a conv layer into several conv layers with smaller kernel size, hence leads to a growing model depth with less parameters.
    \item Compression operator $\delta_3$: \textit{channel-wise scaling}  techniques (\eg channel-level pruning~\cite{bib:cai2019once} and channel-wise architecture noise injection~\cite{bib:yang2020progressive}) can tune variable operator-variant sampling.
    \item Compression operator $\delta_4$: \textit{depth scaling} techniques (\eg depth-elastic pruning~\cite{bib:cai2019once}, residual connection~\cite{bib:he2016deep}) can  derive a shallower variant-network from a backbone-network via skipping connections.
\end{itemize}
%

\subsection{Ensemble Training of Self-evolutionary Network}
We put the retraining process ahead in the ensemble training of the self-evolutionary network at design time to get rid of weight retraining during dynamic inference.
Therefore, the self-evolutionary network training is an ensemble of a backbone-net and multiple variant-nets derived by various convolutional compression operators.
%

\subsubsection{Primer on Parameter Recycling}
We refer to the parameter recycling strategy~\cite{bib:wu2018deep, bib:cai2017efficient} to recycle the backbone-net weights and take less search time than those searching from scratch.
%
%
%
The \textit{weight recycling} strategy is conducive to making maximum use of the existing architectures' experience to reduce the time complexity of the searching process.
In particular, we reuse the existing hand-crafted/elaborated high-performance DNN, including architecture and weight, as an initialization point (\ie backbone network). 
And then, we leverage an automated optimizer (\ie search strategy) to only search for the optimal architecture adjustments (\eg widening a certain network, skipping connections) to obtain a promising new model.
%
%
However, these methods challenge the ensemble training of multiple variant models that drift away from the backbone-net's initial configuration.
In detail, the training of a variant's weights will likely interfere/override the weights that have been learned for other variants and thus degrade the overall performance. 
We note that the catastrophic interference problem when multiple variant-nets share parameters is a long-standing problem in itself~\cite{bib:french1999catastrophic, bib:kirkpatrick2017overcoming}.

\subsubsection{Training Strategy}
Our goal with the self-evolutionary network is to integrate with multiple versions of DNN architectures and the corresponding weights introduced by different compression operators.
And the above-mentioned parameter recycling strategy provides great potentialities.
To further avoid the catastrophic interference problem caused by parameter recycling, we present a novel training strategy to consider the parameter transformation and knowledge distillation for preserving the parametric function of multiple variant-nets.
In detail, we first perform the standard back-propagating process to train a high-accuracy backbone-net. 
Afterwards, we respectively leverage the parameter transformation techniques for learning compression operators $\delta_1$ and $\delta_2$, the knowledge distillation techniques for learning compression operators $\delta_3$ and $\delta_4$, and the trainable channel-wise mutation techniques for learning $\delta_3$.
%
\begin{enumerate}

\item \textbf{Parameter transformations} for learning variant-nets derived by compression operator $\delta_1$ and $\delta_2$. 
We consider the function-preserving parameter transformation when recycling parameters. 
It allows us to initialize a new variant-net that is derived by a compression operator to preserve the function of the given backbone-net, but use different parameterization to be further trained to improve the performance~\cite{bib:cai2017efficient}.
We transform the original convolutional parameter and store the extra copy of weight for $\delta_1$ and $\delta_2$.
%
%
And we further set an accuracy target as a threshold, by which the transformed parameters for compression operator-variants will only be fine-tuned when its accuracy is lower than that.
As thus, we need a small number of extra parameters to store the transformed parameters for compression operators $\delta_1$ and $\delta_2$.
And we only access the weights of the deterministically selected compression operator to evolve model weights.

\item \textbf{Knowledge distillation} for learning variant-nets derived by compression operator $\delta_3$ and $\delta_4$. 
We allow each conv layer to choose the depth/channel compression ratio flexibly.
%
And we adopt the knowledge distillation techniques~\cite{ bib:cai2019once} to fine-tune the parameters of compression operators $\delta_3$ and $\delta_4$ with different compression ratios (\eg $20\%, 50\%$). 
So that \systemname can flexibly switch over the different parameter weights of channel-wise and depth-wise scaling operators, and avoid weight interference.
Also, we perform the trainable channel-wise and depth-wise architecture ranking as the weight importance criterion to guide the adaptive layer slimming and scaling.
In particular, we pre-train a self-evolutionary network to evaluate the overall performance (\eg accuracy drop, parameter arithmetic intensity, activation arithmetic intensity, and latency) of different variant-networks that are compressed by different operators. 
And these are used as the prior-based architecture importance ranking to guide the runtime scaling to shrink unimportant layer/channel first, rather than randomly scaling.

\item \textbf{Trainable channel-wise mutation} for training variant-nets derived by compression operator $\delta_3$.
To maintain a good diversity of solutions, we present a novel trainable architecture mutation technique to inject the architecture variance into the compressed network. 
This idea is supported by recent DNN studies, which have verified the dominant effect of model architecture on accuracy compared to model parameters~\cite{bib:yu2019autoslim}.
That is, \systemname can directly use the trainable architecture mutation technique with diverse noise magnitude since the channel importance ranking of the backbone-net is consistent at both design time and runtime. 
Specifically, we inject Gaussian noise to the channel-wise operator's scaling ratio (\ie $\delta_3$), and the noise magnitude is trainable for channel importance ranking.
That is, the more important the channel is, the lower intensity of noise we inject.
This, as we will evaluate in $\S$ \ref{subsec_exp_bench}, plays a nontrivial role for \systemnameposs progressive shortest encoding process of DNN compression configurations and the runtime searching process for boosting the runtime DNN adaptation quality and efficiency.
\end{enumerate}
Besides, to enable the stable ensemble training of multiple variant-nets, we leverage the mini-batch techniques to split the training data into small batches. We normalize the gradient to reduce the interference caused by gradient variance~\cite{bib:li2014efficient}.

%% file: body/online.tex
\section{Runtime Adaptive Compression}
\label{sec:online}
%
%
This section presents how \systemname quickly searches for the most suitable combination of retraining-free compression operators, from a flexible and elite space, to reconfigure the trained self-evolutionary network on-the-fly.
%

\subsection{Flexible and Elite Search Space}
\label{subsec:space}
%
%
%
\subsubsection{Multi-granularity Search Space}
We form \rev{an elite search space, which include a set of coarse-grained compression operators (\eg Fire block~\cite{bib:iandola2016squeezenet}, SVD-based~\cite{bib:wu2018deep}, sparse coding-based~\cite{bib:bhattacharya2016sparsification} factorization), for faster convergence, and the fine-grained compression operators (\eg channel-level and depth-level pruning and channel-wise randomization), for better diversity.}
Consider a convolutional layer that has the total parameters: (input feature map channel size $M$) $\times$ (output feature map channel size $N$) $\times$ (kernel width/height $S_P$)$\times$ (kernel width/height $S_P$), and total activations: $N \times$ $S_A$ (output feature map width/height) $\times$ $S_A$.
Scaling either the input feature map, kernel, channel, or output feature map can shrink the model complexity.
We empirically observe that different scaling dimensions are not independent. 
Firstly, there is no single compression technique that achieves the best application-driven performance (\ie $A$, $T$, $E$, $C$, $S_a$,  and $S_p$). 
It is necessary to combine several compression techniques.
Secondly, as mentioned in $\S$ \ref{sec:related}, few existing compression techniques are retraining-free or dedicated to optimizing the holistic hardware efficiency across various platforms. 
These findings further suggest us to\textit{flexibly coordinate and balance multiple scaling dimensions} by searching for the best combination of compression operators, rather than the single dimension (\eg model pruning). %

\subsubsection{Hardware Efficiency-guided Combination.}
\label{subsec:hardmetric}
%
%
%
%
We argue that the widely used parameters number, MAC amount, or speedup ratio are not good approximations for hardware efficiency, which heavily depends on the memory movement and bandwidth bound. 
For example, Jha \etal~\cite{bib:jha2019ramifications} reported that although SqueezeNet~\cite{bib:iandola2016squeezenet} has $51.8\times$ fewer parameters than AlexNet~\cite{model:alexnet}, it consumes 33$\%$ more energy due to its larger amount of activations and data movement. 
%
%
And we identify that merely cutting down the parameter size may lead to an increase in activation size, which, in turn, increases the memory footprint and energy consumption~\cite{bib:jha2020modeling}.
For example, the recent study~\cite{bib:jha2019ramifications, bib:jha2020modeling} has shown that the energy consumption of CNNs mainly depends on the memory movement, memory reuse, and bandwidth bound.

\begin{figure*}[t]
  \centering
  \includegraphics[width=.9\textwidth]{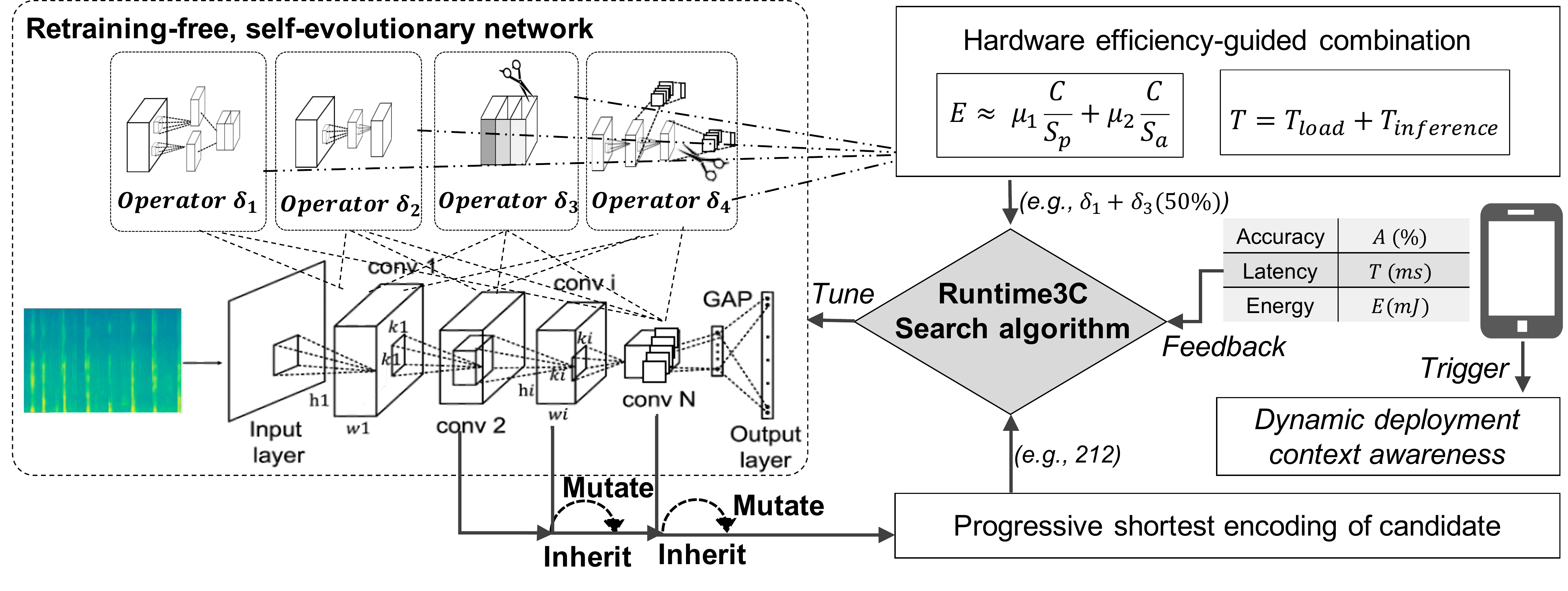}
  \caption{\systemname dynamically shrinks the convolutional compression configurations (3C) of the backbone-net by searching and applying the most suitable combination of retraining-free compression operators using Runtime3C algorithm at runtime.}
  \label{fig:runtime}
\end{figure*}

To this end, we present the controllable hardware-efficiency criteria, \ie arithmetic intensity, to guide the automated combination of compression operators in different layers.
We leverage the arithmetic intensity as a proxy to the degree of reuse of parameters and activations and the energy consumption required for processing inputs, inspired by hardware studies~\cite{bib:jha2019ramifications, bib:jha2020modeling}.
Because the measurement of hardware-relative metrics, especially energy efficiency, is not straightforward.
%
%
%
%
%
%
%
%
Thereby, we present three hardware-efficiency metrics to predict how efficiently arithmetic operation can reuse the data fetched from different levels in the memory hierarchy and how efficiently the arithmetic operation is executed.
\begin{itemize}
    \item Computation/parameter ratio $C/S_p$: is an approximation of the parameter arithmetic intensity; 
    \item Computation/Activation ratio $C/S_a$: \rev{is the proxy of the activation arithmetic intensity};
    \item latency $T$: include the measured inference time $T_{inference}$ of a specialized model, and the time \ie $T_{load}$ for loading parameters and activations for convolution computing on the target mobile device, \ie $T=T_{load}+T_{inference}$.
\end{itemize}
We separately evaluate $C/S_p$ and  $C/S_a$ and then aggregate them together by the aggregation coefficients $\mu_1$ and $ \mu_2$, to better profile the energy efficiency of each candidate compression operator $i$.
\begin{equation}
    E \approx \mu _1 C/S_p + \mu_2 C/S_a
\label{equ_E}
\end{equation}
%
%

Upon these criterions, \systemname automatically selects and combines compression operators for maximizing the aggregated value of $\mu _1 C/S_p + \mu C/S_a$, according to the upper limit of the calculation intensity of the mobile platforms.
And test the real latency $T$ to prevent the exploration of invalid solutions via comparing with the latency budgets.
We empirically set $\mu > \mu$ (\eg $\mu=0.4, \mu=0.6$ as default) since $C/S_a$ contributes more to memory footprint (as benched in $\S$ \ref{subsec_exp_bench}).
%
%
And \systemname discovers some novel combinations for optimizing the underlying data movement. 
For example, we suggest the $\delta_1 + \delta_3$ and  $\delta_2 + \delta_4$ groups (as discussed in $\S$ \ref{sec:experiment}). 
The fine-grained channel-wise scaling operators (\eg $\delta_3$, $\delta_4$) readjust the channel size, MAC amount, and output activation size of the conv layers to smooth out the bandwidth bound problem, which is caused by the coarse-grained operators (\eg $\delta_1$ and $\delta_2$).
The hardware efficiency-guided combination of several compression operators also helps to avoid the blindly explosive combination.

\subsection{Runtime Search Strategy}
\label{subsec:search}
To evolve DNN architecture and weight to an optimal configuration at runtime, we propose the runtime search strategy based on the above flexible and elite search space.


\subsubsection{Progressive Shortest Encoding of Candidate}
Consider a complex self-evolutionary network that contains many combinations of compression operator variants and configurations, systematically and generically choosing the right candidate configurations and encoding them into the representation of a search algorithm is difficult.
%
%
For optimizing DNN compression configurations at runtime, such representations define the potential search space of the problem to be explored.
Given that some candidate configurations do not contribute to the specific performance optimization demand or other candidates can represent some of their information, the shortest actual encoding will benefit the search result (\ie model evolution plans) and overhead.
%

\begin{figure}[t]
  \centering
  \subfloat[Classic binary encoding.]{
  \includegraphics[height=0.24\textwidth]{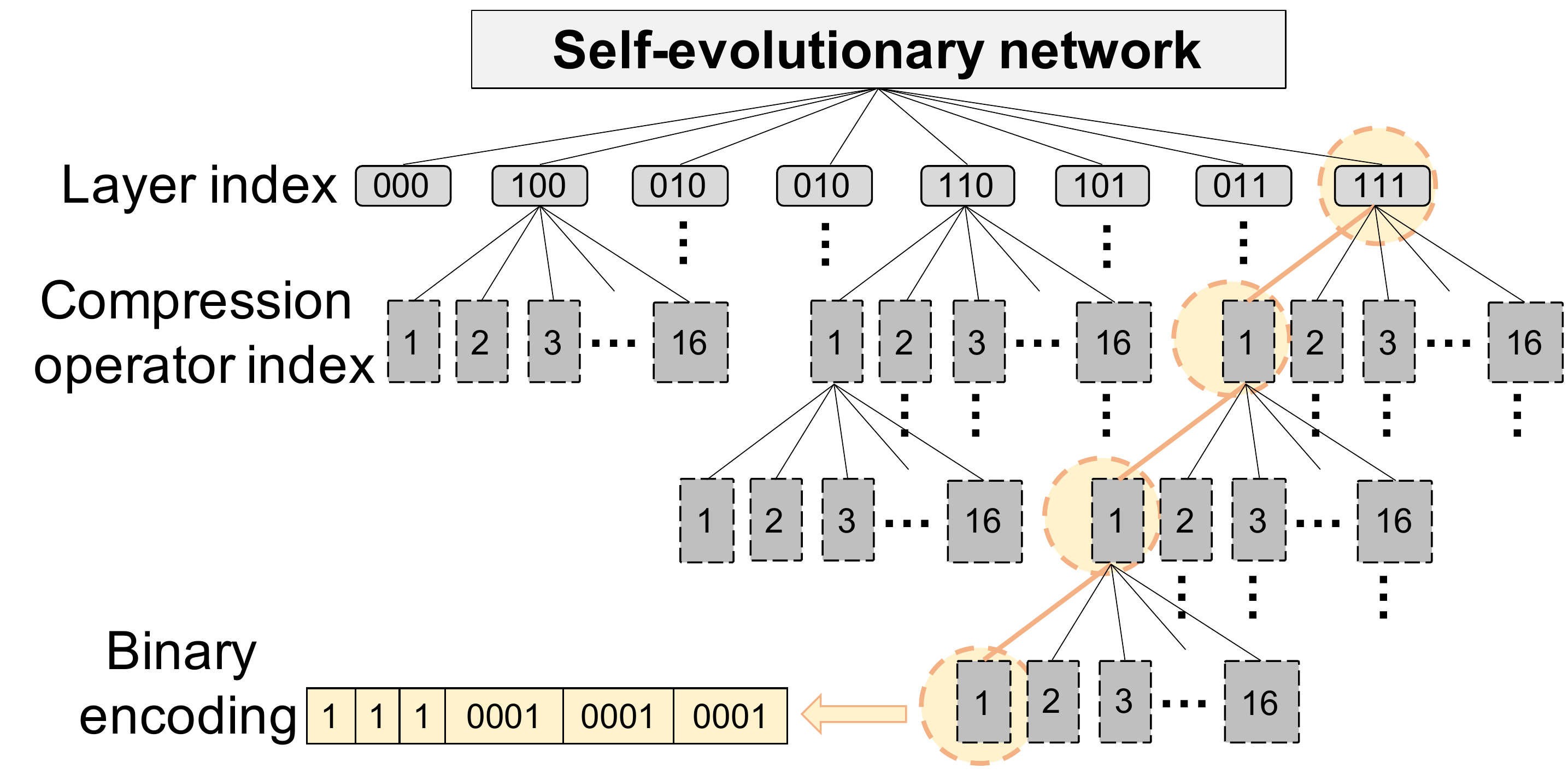}}
\hfill
   \subfloat[Progressive shortest encoding.]{
   \includegraphics[height=0.24\textwidth]{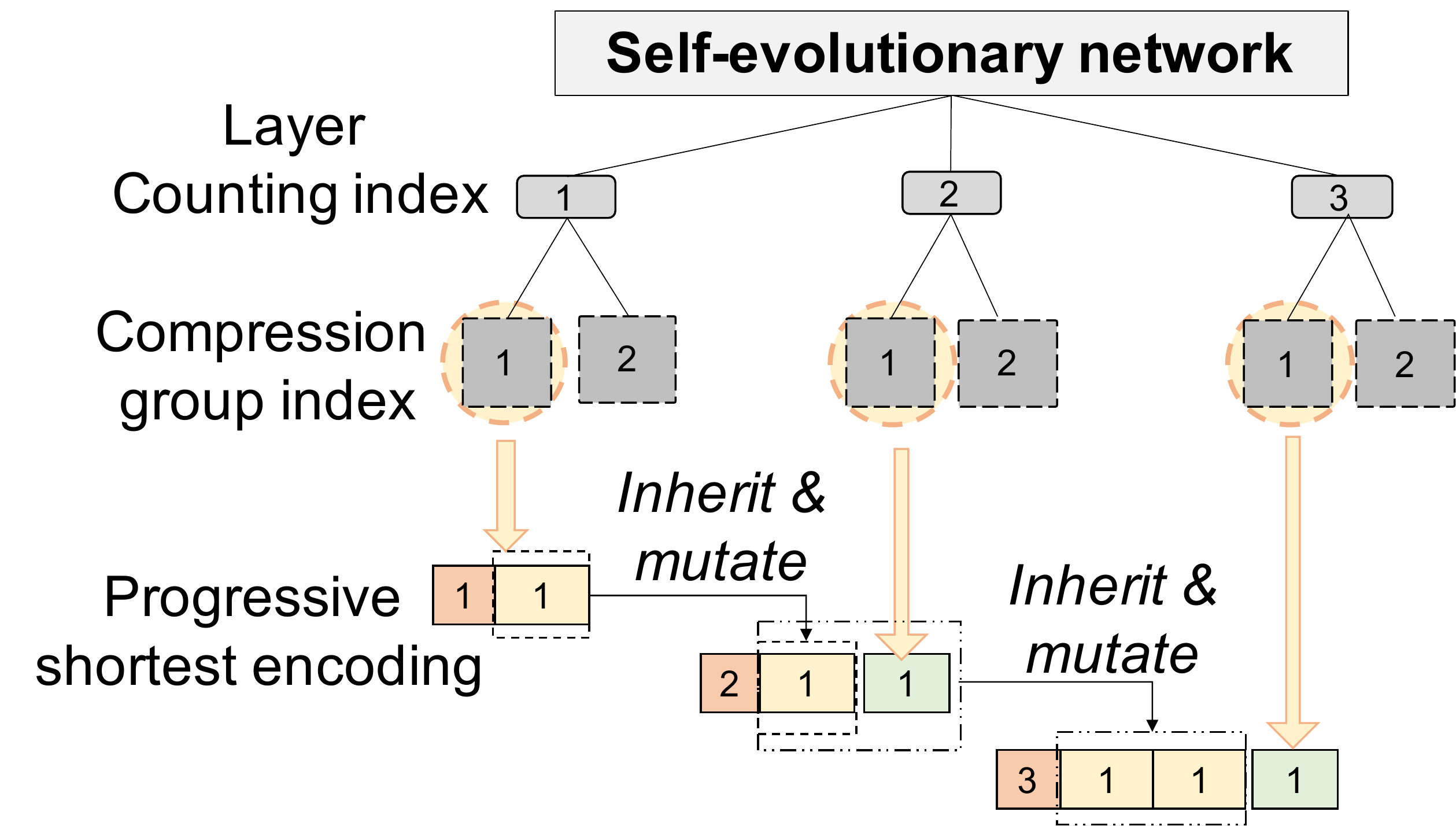}}
\caption{\systemname adopts the progressive shortest encoding of compression operator-variant configurations via a layer-dependent manner, \ie (b), to reduce the search complexity, compared to the classic binary encoding of that, \ie (a).
}
\label{fig_encoding}
\end{figure}

As shown in Figure \ref{fig_encoding}(a), the classic binary encoding of all compression operator configurations across all layers in a binary format is redundant.
Specifically, given a backbone network with $N$ conv layers to be selectively compressed. 
Take $N=3$ as an example. A classic binary encoding method needs $3$ bit to record whether a specific layer participates in compression or not. Other $3*4$ bits (four bits to represent $4^2$ selective operators per layer). 
In this way, the encoding length is $N+MN=(M+1)N$ when we have $M$ optional compression operator. And the search space derived by this encoding diversity is $2^N \times{M^N}$, \ie \textit{$O(M^{N})$}. Furthermore, it will increase exponentially as the number of optional compression operators increase.

To better represent the fundamental search space, we propose the \textit{progressive shortest encoding} of compression operator configurations via a layer-dependent manner.
As we will show in $\S$ \ref{subsubsec:exp_encoding}, it improves the search efficiency by one order of magnitude, compared to the classic binary encoding.
As shown in Figure \ref{fig_encoding}(b), we use $N$ digits to record the count of layers that have been compressed.
The first digit represents the compressed layer count, and the next length-variable few digits record the selected compression operator index of each layer. 
For example, the value $1$ of the first digit means that only the first conv layer is compressed on-demand.
Thereby, only one additional digit is needed to record the compression operator index (\ie, $1$) for it.
Afterward, \systemname inherits the above $2$-digit string and inject channel-wise variance to mutate the inherited survival $2$-digit encoding string.
We refer the channel-wise variance mutation process in $\S$ \ref{subsec:search}.
And then, we turn to the second adaptable conv layer. If the first digit of compressed layer count is updated to 2, we append one more digit indicating the selected compression operator index to the survival $2$-digit encoding string.
Thus, the encoding length progressively increases from 2 to ($N+1$), and the complexity of the search space is reduced to \textit{$O(N^2)$}.
The progressive shortest encoding of the candidate is conducive to the flexibility of  \systemname and prevents unnecessary exploration.

\subsubsection{Runtime3C Search Algorithm}
This subsection presents the Runtime3C search algorithm, a Pareto optimal decision-based searching algorithm, to pick a sole optimal solution from the search space at runtime.
%
%
To the best of our knowledge, many widely used universal search algorithms (\eg evolutionary algorithms) are not designed to optimize the runtime adaptive compression problem or handle dependency constraints of multiple DNN performance.
We heuristically regard the selection of compression operators for each layer as a \textit{single-layer optimization subproblem} in a collaborative manner to derive the most suitable solution quickly and effectively.

\begin{algorithm}[t]
 \LinesNumbered
 \KwIn{(a)\textbf{Deployment context}: dynamic context constraints $\theta_{A_{loss}}, T_{bgt}, S_{bgt}$; relative importance $\lambda_1$, $\lambda_2$.  \;
 (b) \textbf{DNN}: a trained self-evolutionary network comprising of a backbone-net $\Omega$ and multiple retraining-free compression operator-variants $\mathcal{\delta}$,  $\delta \in \mathrm{\Delta} $.}
 \KwOut{A reconfigured DNN model $\Omega(\delta, w_{\delta}) $ }
 Transforming search space from $(\delta \in \mathrm{\Delta})$ to hardware-efficient group $(group (\delta) \in \mathrm{\Delta '})$   \;
\While { (layer $i$ is conv type) $\& \&$ (layer $i$ is to be compressed)}{
    Inherit 3C configurations from layer (i-1) $\mathcal{E} \{ \Omega , group(\delta_(i-1)) \}$ \;
    Select 2 candidates $group(\delta_j), group(\delta_k)$ from the Pareto front of the valid space $\mathrm{\Delta ' }$\;
    Mutate and augment 2 candidates to 6 candidates using the trained channel-wise variances $\epsilon$ \;
    Select the Pareto-optimal candidate (\ie min $A_{loss}$ while max $E$) as the survival candidate $group(\delta_i)$  \;
    Evolve weights $w_{\delta_i}$ via parameter transformation  \;
    Encode candidates $\mathcal{E} \{ \Omega , group(\delta_i) \}$ \;
    Forward DNN $\Omega(group(\delta_i))$ to measure $A$, $T$, $S_p$, $S_a$, $C$, $E$ of the overall model $\Omega(\delta_i)$ \;
    Judge whether the DNN performance satisfy constraints of the current deployment context\;
    \If{context constraints satisfied}{ Searching stops;}
    layer ++;
}       
*Note: we start exploring compression operator configurations from the second conv layer by default to preserve more input details.
 \caption{Pareto decision-based Runtime search algorithm for convolutional compression operator configurations (Runtime3C)}
 \label{alg_search}
\end{algorithm}


As shown in Algorithm~\ref{alg_search}, each subproblem at layer $i$ is to search the optimal group of compression operators for optimizing the overall performance of the entire DNN.
Starting from the second conv layer by default, \systemname selects two candidate solutions at layer $i$ from the Pareto front of the selectable compression operator groups for optimizing the accuracy and energy efficiency of the entire model  (line 2).
In detail, the picked two candidate solutions are the best two compromises in $\lambda_1 log(A_{loss})$ \vs $\lambda_2 log(E)$, from the Pareto front within the valid search space, \ie $A_{loss}>5\%$.
\rev{Here, we leverage the ranking of the pre-tested accuracy and energy cost of the DNNs to establish the Pareto front. And the accuracy ranking derived by historical results is consistent with the ranking of the actual accuracy of these DNNs measured on mobile devices. 
}
We then mutate and augment candidates from two to six by injecting the channel-wise variance to the candidate configurations. The trained architecture importance is a criterion for Gaussian noise injection. 
This process can improve the diversity of subproblem solution as well as the performance of the global solution, inspired by the genetic algorithm in the adaptive software engineering~\cite{bib:chen2018femosaa}.
%
%
We choose the best candidate as the survival subproblem solution for compressing layer $i$ (line 6). 
Afterward, the $i-$th layer's survival solution is used to reconfigure the $i-$layer and becomes the initial station of the subproblem at $(i+1)$-th layer.
We fix the selected compression configurations for $i-$th layer and repeat the above-searching steps (line $3 \sim 9$) to specialize the optimal compression strategies for the $(i+1)$-th layer.
Once the model satisfies the dynamic constraints in latency $T_{bgt}(t)$ and memory $S_bgt(t)$ at time $t$, the subproblem expansion stops (line 12).
And finally, it outputs the global compression configuration solution.

%


%% file: body/experiment.tex
\section{Evaluation}
\label{sec:experiment}
This section presents the evaluation of \systemname over different mobile applications on diverse mobile and embedded platforms with dynamic deployment context. We compare \systemname against ten alternative methods reported in the state-of-the-art literature.

\begin{table}[t]
\scriptsize
\caption{Summary of the applications and corresponding datasets for evaluating \systemname}
\begin{tabular}{|l|l|l|l|}
\hline
\textbf{\textbf{No.}} & \textbf{Target task (utility label)} & \textbf{Dataset} & \textbf{Description} \\ \hline
$D_{1}$ & Image ($10$ classes) & \rev{CIFAR-100}\cite{data:cifar} & $60,000$ images \\ \hline
$D_{2}$ & Image ($5$ classes) & ImageNet\cite{data:imagenet} & $65,000$ images \\ \hline
$D_{3} $ & Acoustic event ($9$ classes) & UbiSound\cite{bib:sicong2017ubiear} & $7,500$ audio clips \\ \hline
$D_{4}$ & Human activity ($7$ classes) & Har\cite{data:Har} & $10,000$ records of accelerometer and gyroscope \\ \hline
$D_{5}$ & Driver behavior ($10 $ classes) & StateFarm\cite{data:statefarm} & $22,424$ images \\ \hline
\end{tabular}
\label{tb:sum_tasks}
\end{table}

\subsection{Experiment Setup}
\label{subsec_expset}
We first present the settings for our evaluation.

\textbf{System Implementation}. 
We implement \systemnameposs offline block with TensorFlow \cite{url:tensorflow} in Python on the server side to train the self-evolutionary network (see $\S$ \ref{sec:offline}).
And we realize the \systemnameposs online blocks on the mobile and embedded platforms to adjust the DNN configurations on the fly for better inference performance.
The self-evolutionary network (\ie a backbone-net and multiple variant compression operators), generated by \systemnameposs offline component, is then loaded into the target platform.
%
%
To further reduce the memory access cost, we load DNN parameters from L2-Cache memory.

\textbf{Evaluation Applications/Datasets}. 
We use five commonly used mobile applications/datasets to evaluate  \systemnameposs performance as elaborated in Table~\ref{tb:sum_tasks}.
Specifically, we test \systemname for mobile image classification (D1: Cifar100~\cite{data:cifar100}, D2: ImageNet~\cite{data:imagenet}), 
mobile acoustic event awareness (D3: UbiSound~\cite{bib:sicong2017ubiear}), 
mobile human activity sensing (D4: Har~\cite{data:Har}), and mobile driver behavior prediction (D5: StateFarm~\cite{data:statefarm}).

\textbf{Mobile Platforms with Dynamic Context Settings}. We evaluate \systemname on \rev{three categories of commonly used mobile and embedded platforms}, including one personal smartphones, \ie Xiaomi RedMi 3S (device1), one embedded development board, \ie raspberry Pi 4B (device3), and one mobile robot platform \ie NVIDIA Jetbot (device4) loaded with the mobile development board.
They are equipped with diverse processors, storage and battery capacity.
The dynamic context is formulated by the time-varying latency budget $T_{bgt}(t)$, storage budget $S_{p}(t)$, and the relative importance coefficient of accuracy and energy efficiency objectives.
%

\textbf{Comparison Baselines}. We employ three categories of DNN specialization baselines to evaluate $\systemname$.
%
%
%
%
The detailed settings of ten baselines from three categories are as below. Firstly, the hand-crafted compression baselines relay on manual design to realize efficient DNN compression. \rev{They provide the high standard for \systemname to tune the specialized DNNs' performance tradeoff between accuracy, latency, and resource efficiency.}
\begin{itemize}
    \item Fire~\cite{bib:iandola2016squeezenet} presented in SqueezeNet reduces filter size and decreases input channels using squeeze layers.
    \item MobileNetV2~\cite{bib:sandler2018mobilenetv2} replaces the traditional convolutional operation by an inverted residual with the linear bottleneck to expand module to high dimension and then filter with a depth-wise convolution. 
    %
    \item SVD-based convolutional decomposition technique~\cite{bib:lane2016deepx} introduces an extra conv layer between $conv_i$ and $conv_{(i+1)}$ using the singular value decomposition (SVD) based parameter matrix factorization. The number of neurons $k$ in the inserted layer is set according to the dynamic neuron numbers $m$ in $conv_i$, \ie $k=m/12$.
    \item Sparse coding-based convolutional decomposition technique~\cite{bib:bhattacharya2016sparsification} insert a conv layer between $conv_i$ and $conv_{(i+1)}$ using the sparse coding-based parameter matrix factorization. The k-basis dictionary is dynamically determined by the neuron number $m$ in $conv_i$, \ie $k=m/6$.
    \end{itemize}
    Secondly, the on-demand DNN compression baseline methods adopt a trainable optimizer to automatically find the most suitable DNN compression strategies for various mobile platforms.\rev{These baselines provide a strict benchmark against which we can validate that both searching and retraining costs are bottleneck limitations for the runtime adaptation demands. }
    \begin{itemize}

    \item AdaDeep~\cite{bib:liu2020adadeep} automatically selects and combines compression techniques to generate a specialized DNN that balance accuracy and resource constraints.
    \item ProxylessNAS~\cite{bib:cai2018proxylessnas} directly learns architectures without any proxy while still allowing a large candidate set and removing the restriction of repeating blocks.
    
    \item Once-for-all(OFA)~\cite{bib:cai2019once} obtains a specialized sub-network by selecting from the once-for-all network that supports diverse architectural settings without additional training.
    \end{itemize}
Thirdly, the runtime adaptive DNN compression requires to search for the most suitable combination of retraining-free compression techniques quickly, we select two baseline optimization methods to compare with \systemname.  \rev{Here, the baseline optimizers represent two intuitive searching ideas for the runtime adaptive compression of DNN configurations.}
    \begin{itemize}
    \item Exhaustive optimizer tests all combinations of compression operators' performance on the validation and then selects the one variety with the best tradeoff based on the fixed performance ranking. \rev{And then it fixes the compression operators and only scale down the compression operators' hyperparameters, \ie compression ratio, to satisfy the dynamic resource budgets.}
    \item Greedy optimizer selects the best compression operator layer-by-layer that obtains the best tradeoff between accuracy and parameter size, in which the relative importance is equally set to a fixed value of $0.5$.
    \item \systemname selects and applies the most suitable combination of compression operators into the self-evolutionary backbone network for accuracy and resource efficiency tradeoff.
\end{itemize}

\begin{table}[t]
\scriptsize
\caption{Performance comparison of \systemname with three categories of baselines on Raspberry Pi 4B (device 2) using CIFAR-100 ($D_1$). The backbone network includes 5 conv layers, and 1 GAP layer. } 
\begin{threeparttable} 
\begin{tabular}{|c|c|c|c|c|c|c|c|c|c|c|}
\hline
\multirow{2}{*}{\textbf{Baselines}} & \multirow{2}{*}{\textbf{\begin{tabular}[c]{@{}c@{}}DNN compression\\ techniques\end{tabular}}} & \multicolumn{5}{c|}{\textbf{Performance of specialized DNN}} & \multicolumn{4}{c|}{\textbf{Performance of DNN specialization scheme}} \\ \cline{3-11} 
 &  & \textbf{$A$ ($\%$) \tnote{*1}} & \textbf{$T$ (ms)} & \textbf{$ C/S_p$} & $ C/S_a$ & \textbf{$En$(mJ)} & \textbf{\begin{tabular}[c]{@{}c@{}}Search \\ cost \end{tabular}} & \textbf{\begin{tabular}[c]{@{}c@{}}Retraining \\ cost (hours)\end{tabular}} & \textbf{\begin{tabular}[c]{@{}c@{}}Scale \\ down\end{tabular}} & \textbf{\begin{tabular}[c]{@{}c@{}}Scale \\ up\end{tabular}} \\ \hline
\multirow{4}{*}{\textbf{\begin{tabular}[c]{@{}c@{}}Stand-alone\\ compression\end{tabular}}} & Fire~\cite{bib:iandola2016squeezenet} & 72.3 & 24.7 & 81.2 & 394.7 & 3.1 & 0 & 1.5N \tnote{*2}& fix &  \\ \cline{2-11} 
 & MobileNetV2~\cite{bib:sandler2018mobilenetv2} & 72.6 & 48.1 & 84.3 & 128.4 & 5.2 & 0 & 1.8N \tnote{*2}& fix & --- \\ \cline{2-11} 
 & SVD decomposition~\cite{bib:lane2016deepx} & 71.2 & 21.7 & 68.6 & 165.8 & 4.8 & 0 & 2.3N \tnote{*2}& scalable & --- \\ \cline{2-11} 
 & \begin{tabular}[c]{@{}c@{}}Sparse coding \\ decomposition~\cite{bib:bhattacharya2016sparsification} \end{tabular} & 72.9 & 22.3 & 69.8 & 195.2 & 4.6 & 0 & 2.3N\tnote{*2} & scalable & --- \\ \hline
\multirow{3}{*}{\textbf{\begin{tabular}[c]{@{}c@{}}On-demand \\ compression\end{tabular}}} & AdaDeep~\cite{bib:liu2020adadeep} & 73.5 & 21.9 & 78.3 & 264.6 & 3.5 & 18N $hours$ \tnote{*2}& 38N \tnote{*2}& scalable & --- \\ \cline{2-11} 
 & ProxylessNAS~\cite{bib:cai2018proxylessnas} & 74.2 & 49.5 & 121.3 & 232.1 & 3.8 & 196N $hours$ \tnote{2}& 29N\tnote{*2}& scalable & --- \\ \cline{2-11} 
 & OFA~\cite{bib:cai2019once} & 71.4 & 51.2 & 123.4 & 257.3 & 3.1 & 41 $hours$ & 0 & scalable & scalable \\ \hline
\multirow{3}{*}{\textbf{\begin{tabular}[c]{@{}c@{}}Runtime\\ adaptive \\ compression\end{tabular}}} & Exhaustive optimizer & 58.3 & 21.1 & 81.2 & 283.2 & 2.9 & 0 & 0 & --- & --- \\ \cline{2-11} 
 & Greedy optimizer & 65.3 & 16.7 & 83.5 & 298.4 & 3.1 & 25 $ms$ & 0 & --- & --- \\ \cline{2-11}
 & \systemname & 74.1 & 15.6 & 158.9 & 358.7 & 1.9 & 3.8 $ms$ & 0 & scalable & scalable \\ \hline
\end{tabular}
\begin{tablenotes}    
        \footnotesize    
        \item[*1] We test the average DNN accuracy at three dynamic moments.
        \item[*2] The $N$ in search cost and retraining cost columns shows that the cost is linear to the number of deployment contexts.         
      \end{tablenotes}    
\end{threeparttable} 
\label{tb_compare}
\vspace{-2mm}
\end{table}

\subsection{Performance Comparison}
\label{subsec_compare}
We evaluate \systemname in terms of the specialized DNNs' running performance (\ie accuracy $A$, amount of MACs $C$, parameter arithmetic intensity $C/S_p$, activation arithmetic intensity $C/S_a$, and energy consumption $EC$) and the specialization methods' all-around performance (\ie search cost, retraining cost, and scaling flexibility).
As shown in table \ref{tb_compare}, we compare \systemnameposs performance with ten baselines.
In this thread of experiments, we leverage the same mobile sensing task (\ie image recognition using CIFAR-100 ($D_1$) datasets) and target embedded platform (\ie Raspberry Pi 4B) for six state-of-the-art baselines and \systemname for a fair comparison.
We adopt different baseline methods to specialize the DNN architectures and weights for optimizing accuracy and resource efficiency objectives with dynamically specified constraints (see $\S$ Equ. \ref{equ_opt}).
Here, the relative importance coefficients (\ie $\lambda_1$ and $\lambda_2$) are dynamically determined by the remaining battery percentage $E_{remaining}$ of the target platform, \ie $\lambda_2 = max \{0.3, E_{remaining}\}$, and $\lambda_1 = 1- \lambda_2$ .
Afterwards, we test the specialized DNN's running performance on a Raspberry Pi 4B platform.
To mitigate the effect of noise and increase the robustness of performance measurements, we repeat the steps mentioned above five times and take an average over them.

\begin{figure*}[tb]
  \centering
  \includegraphics[width=.99\textwidth]{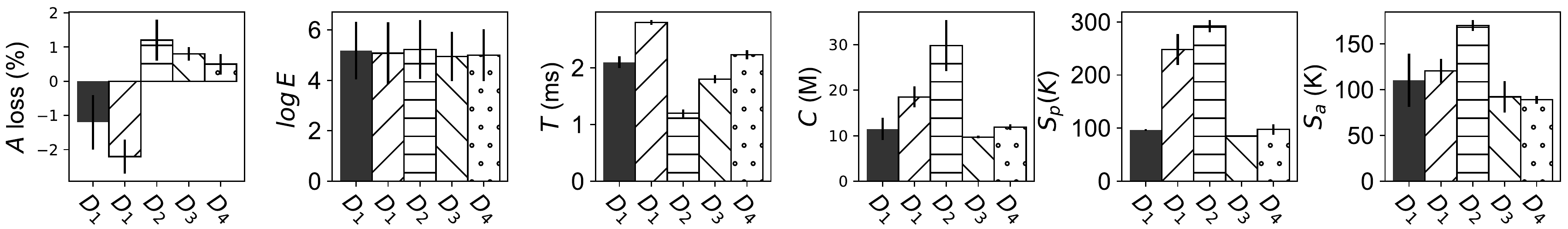}
  \caption{Overall performance of \systemname evaluated on five tasks ($D_1, D_2, D_3, D_4, D_5$) using a Raspberry Pi 4B platform (Device 3). The energy efficiency is measured by $E=0.4* \frac{C}{S_p} + 0.6*\frac{C}{S_a}$, and normalized by $\log(.)$.}
  \label{fig_exp_tasks}
  \vspace{-3mm}
\end{figure*}

\begin{table}[tb]
\scriptsize
\caption{Performance of \systemname evaluated on different tasks/datasets, compared to the corresponding DNNs compressed by depthwise convolutional decomposition, \ie MobileNet.}
\begin{tabular}{|c|c|c|c|c|c|c|c|}
\hline
\multirow{2}{*}{\textbf{\begin{tabular}[c]{@{}c@{}}Mobile  Taks\end{tabular}}} & \multirow{2}{*}{\textbf{\begin{tabular}[c]{@{}c@{}}Compression\\ operator configurations\end{tabular}}} & \multicolumn{6}{c|}{\textbf{Compared to the performance of MobileNet network}} \\ \cline{3-8} 
 &  & \textbf{A loss} & \textbf{E} & \textbf{T} & \textbf{C} & \textbf{Sp} & \textbf{Sa} \\ \hline
\textbf{CIFAR-100($D_1$)} & $ \delta_1+\delta_3(50\%)$ & -2.1\% & 2.5 $\times{}$ & 1.2$\times{}$ & 5.6 $\times{}$& 2.8$\times{}$ & 1.2$\times{}$ \\ \hline
\textbf{ImageNet($D_2$)} & $\delta_2+\delta_4(1)$ & -0.9\% & 8.9$\times{}$ & 1.3$\times{}$ & 8.6$\times{}$ & 5.2$\times{}$ & 1.9$\times{}$ \\ \hline
\textbf{UbiSound($D_3$)} & $\delta_2+\delta_3(75\%)$ & 1.3\% & 15.2$\times{}$ & 1.1$\times{}$ & 4.3$\times{}$ & 3.8$\times{}$ & 1.2 $\times{}$\\ \hline
\textbf{Har($D_4$)} & $\delta\_1+\delta_4(1)$ & -0.3\% & 2.1$\times{}$ & 0.8 $\times{}$& 9.2$\times{}$ & 7.1 $\times{}$& 1.3$\times{}$ \\ \hline
\textbf{StateFarm($D_5$)} & $ \delta_2+ \delta_3(55\%)$ & 0.2\% & 5.9$\times{}$ & 0.7$\times{}$ & 5.6$\times{}$ & 4.3$\times{}$ & 1.6$\times{}$ \\ \hline
\end{tabular}
\vspace{-2mm}
\end{table}

\textbf{Performance comparison}. Table \ref{tb_compare} summarizes the performance comparison between ten baseline methods and \systemname.
First, \systemname achieves the best overall performance in terms of accuracy $A$, MAC amount $C$, parameter arithmetic intensity $C/S_p$, activation arithmetic intensity $C/S_a$, and energy consumption $En$, while incurring negligible accuracy loss, compared to the DNNs specialized by other baseline methods.
The \systemname reduces the model inference latency to $15.6 $ms, the energy consumption to $1.9 mJ$. And it increases the parameter arithmetic intensity $C/S_p$ to $158.9$, the activation arithmetic intensity $C/S_a$ to $358.7$.
Notably, \systemname generates DNN to get the largest activation arithmetic intensity $C/S_a$ and second-largest parameter arithmetic intensity $C_a$.
Compared to the parameter size, the influence of activation arithmetic intensity upon energy consumption is equally or even more critical. 
The DNN specialized by the hand-crafted Fire, MobileNetV2, SVD, and sparse coding techniques consumes energy by $3.1 mJ$, $5.2mJ$, $4.8mJ$, and $4.6mJ$, respectively.
\rev{The accuracy of exhaustive optimizer is much lower than the proposed design, since it shows low accuracy when it fixes the compression operator categories and only over-compresses their hyperparameters. This outcome demonstrates that the reselection of different compression operators are necessary.} 
The specialized DNN's accuracy achieved by \systemname is at least as good as ProxylessNAS, and sometimes even better than the hand-crafted compression techniques.
%
Second, the \systemnameposs specialization scheme is the most efficient in reducing the searching cost and retraining cost.
The adaptive compression baselines involve a high overhead in retraining.
For example, AdaDeep requires an average of $19 \sim 38$ hours for retraining (\eg retraining the deep reinforcement learning model-based optimizer) offline on the GPU platform for each adjustment of compression strategies.
\rev{AdaDeep and ProxylessNAS need $18N$ and $196N$ hours, respectively, to search from the candidate configurations, which increases linearly with the number of dynamic contexts.}
Although OFA and \systemname do not need retaining.
OFA needs $41 hours$ to search per adaptation, while \systemname only needs $3.8 ms$ to do that.
\rev{This is because \systemname leverages the elite compression operator space, rather than the basic kernel size or channel number space in OFA, to avoid the redundant search exploration.}

\textbf{Summary}. \systemname outperforms the other ten baselines in terms of the DNN performance tradeoff between accuracy, latency, arithmetic intensity of parameters and activations, and energy consumption. Meanwhile, it incurs the modest searching cost without retraining, making it ideal for runtime adaptive DNN compression.

\subsection{\systemnameposs Performance over Different Tasks}
\label{seubsec_task}
To illustrate the \systemnameposs performance over different tasks, we evaluate it using all the five applications/datasets (see $\S$ \ref{subsec_expset}) on a Raspberry Pi 4B platforms (Device 3) which is powered by a mobile $3800mAh$ battery. \systemname dynamically detects the platform's remaining battery and sets the  coefficients between accuracy and energy efficiency in Equ. \ref{equ_opt} according to the percentage of remaining power $E_{remaining}$, \ie $\lambda_2 = max \{ 0.3, 1-E_{remaining} \}$. In addition, we specify the storage budget as 2MB that is capacity of the L2-Cache. We set a accuracy loss threshold to be $0.5, 0.3, 0.6, 0.5$ for image classification tasks ($D_1$, $D_2$), sound sensing (\ie $D_3$), human activity prediction task ($D_4$), and driver behavior recognition task ($D_5$), respectively. And assume the latency sensitivity as the latency budget of $20 ms, 10ms$, $30ms, 20ms$ for $D_1 \sim D_5$.

\begin{figure}[t]
  \centering
  \subfloat[RedMi 3S smartphone]{
  \includegraphics[height=0.25\textwidth]{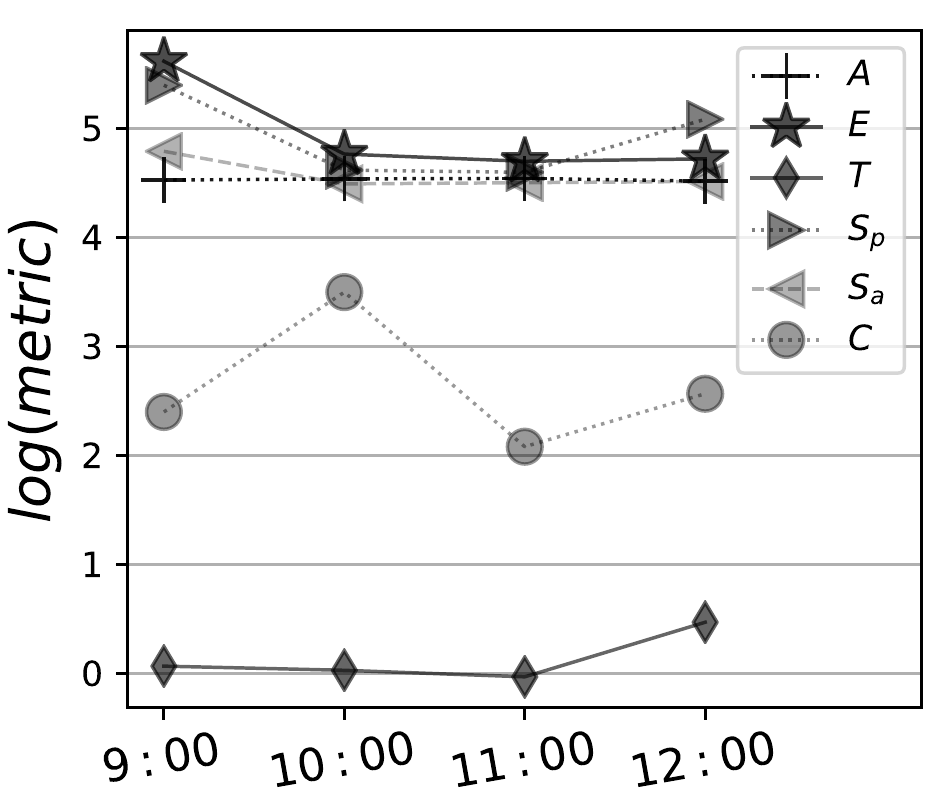}}
  \hfill
  \subfloat[Raspberry Pi 4B]{
  \includegraphics[height=0.25\textwidth]{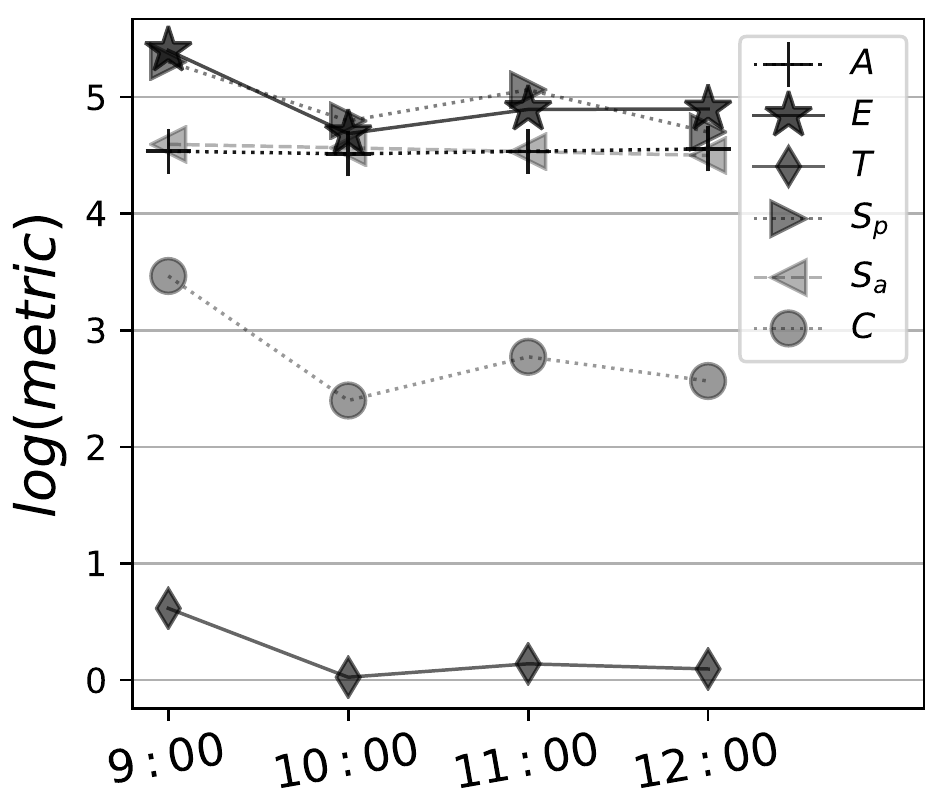}}
    \hfill
  \subfloat[NVIDIA Jetbot]{
  \includegraphics[height=0.25\textwidth]{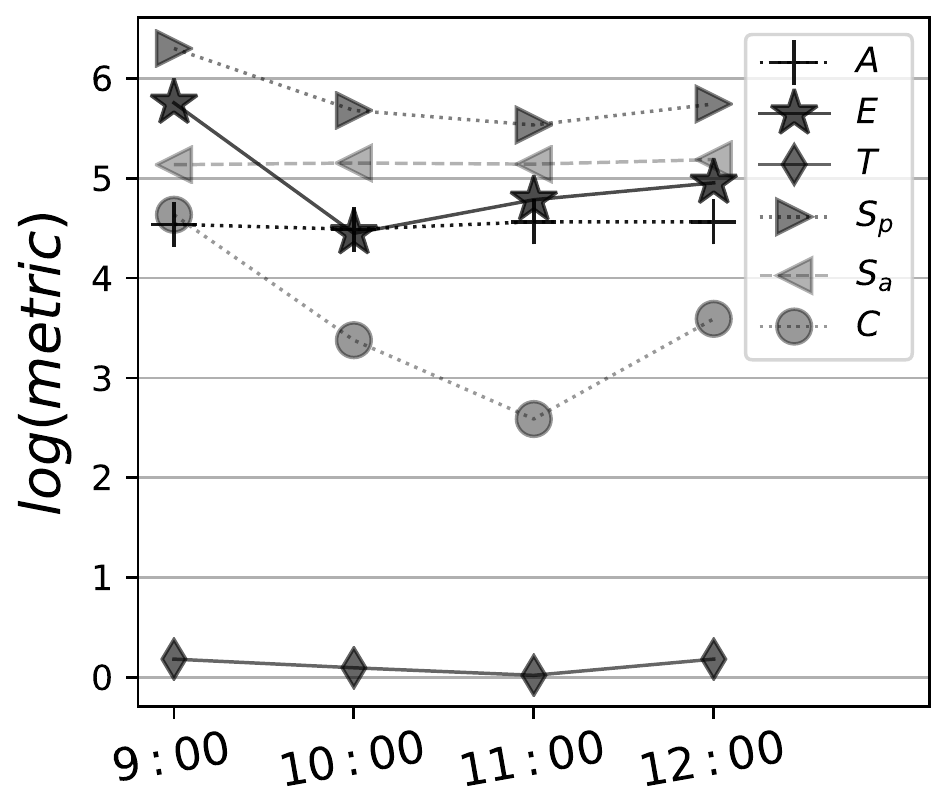}}
\caption{Performance of \systemname evaluated on three platforms for a sound recognition application (D3). And the performance metrics $A, E, T, S_p, S_a, C$ are normalized by $log(.)$.
}
\label{fig_exp_devices}
\vspace{-1mm}
\end{figure}

\begin{table}[t]
\scriptsize
\caption{We test \systemnameposs performance across three platforms on four moments with dynamic contexts.}
\begin{tabular}{|c|c|c|c|l|c|c|c|c|c|}
\cline{1-4} \cline{6-10}
\multicolumn{4}{|c|}{\textbf{Diverse platform}} & \multirow{5}{*}{} & \multicolumn{5}{c|}{\textbf{Dynamic context}} \\ \cline{1-4} \cline{6-10} 
\textbf{Device} & \textbf{Processor} & \textbf{L2-Cache} & \textbf{Battery} &  & \textbf{Time} & \textbf{$9:00am$} & \textbf{$10:00am$} & \textbf{$11:00am$} & \textbf{$12:00 noon$} \\ \cline{1-4} \cline{6-10} 
\textbf{Redmi 3S smartphone} & Qualcomm B21 & 2MB & 4100mAh &  & \textbf{Remaining battery} & $86\%$ & $78\%$ & $72\%$ & $61\%$ \\ \cline{1-4} \cline{6-10} 
\textbf{Raspberry Pi 4B} & Cortex-A72 & 2MB & 3800mAh &  & \textbf{Avaliable cache} & 2MB & 1.6MB & 1.5MB & 1.7MB \\ \cline{1-4} \cline{6-10} 
\textbf{NVIDIA Jetbot} & Cortex-A57 & 2MB & 7200mAh &  & \textbf{Inference require} & 2 times & 1 time & 2 times & 1 time \\ \cline{1-4} \cline{6-10} 
\end{tabular}
\label{tb_exp_device_context}
\vspace{-1mm}
\end{table}

\textbf{Performance}. Figure \ref{fig_exp_tasks} compares the performance of the DNN configurations specialized by \systemname on five different tasks in terms of user experience metrics (\ie inference accuracy $A$, energy efficiency $E$, and inference latency $T$) and direct DNN metrics (\ie computation $C$, parameter size$S_p$, activation size $S_a$).
And we compute the mean and standard deviation of the running performance of the DNN specialized in five dynamic moments, at which the percentage of remaining battery is $0.85$, $0.75$, $0.62$, $0.52$, and $0.38$, respectively. These affect the tradeoff demands on objectives. 
The storage budget $S_{bgt}$ for parameters dynamically depends on the available Cache capacity. 
We simulate the unpredictable resource contention by other software using the randomization noise $\sigma$ injection to Cache's available capacity, \ie $(2-\sigma)$MB. 
For different tasks, datsets, and deployment contexts, \systemname selects the various combinations of compression operators to scale up/down the model configurations to optimize and balance multiple performances. 
It achieves the inference latency $1.2 \sim 2.8$,  the parameter arithmetic intensity $106 \sim 229$, activation arithmetic intensity $143 \sim 220$, with a negligible accuracy loss ($\leq 0.5\%$ ) or even accuracy improvement ($\leq 2.2\%$).  

\textbf{Summary}. For different tasks with diverse backbone model shapes and various sensitivity to accuracy loss and latency, the DNN specialized by \systemname varies. As for the same task, the DNN's compression configurations founded by \systemname also differ according to the dynamic deployment context.

\subsection{\systemnameposs Performance across Diverse and Dynamic Deployment Contexts}
\label{subsec_plat}
In this experiment, we compare the \systemnameposs performance for mobile sound sensing application ($D_3$), tested in three different platforms.
We adopt the same self-evolutionary network comprising of the same backbone-net and some optional compression operator-variants.
Different platforms have different resource characteristics, which are further affected by dynamic deployment contexts. 
In particular, we adopt the RedMi 3S smartphone equipped with Qualcomm B21 processor, $2MB$ L2-Cache, and $4100mAh$ battery; the Raspberry Pi 4B with $2MB$ L2-Cache, and $3800mAh$ battery; and the NVIDIA Jetbot with quad-core ARM Cortex-A57 processor, $2MB$ L2-Cache, and $7200mAh$ battery.
We adopt the similar dynamic deployment context settings with $\S$ \ref{seubsec_task}.

\begin{figure}[t]
  \centering
  \subfloat[]{
  \includegraphics[height=0.21\textwidth]{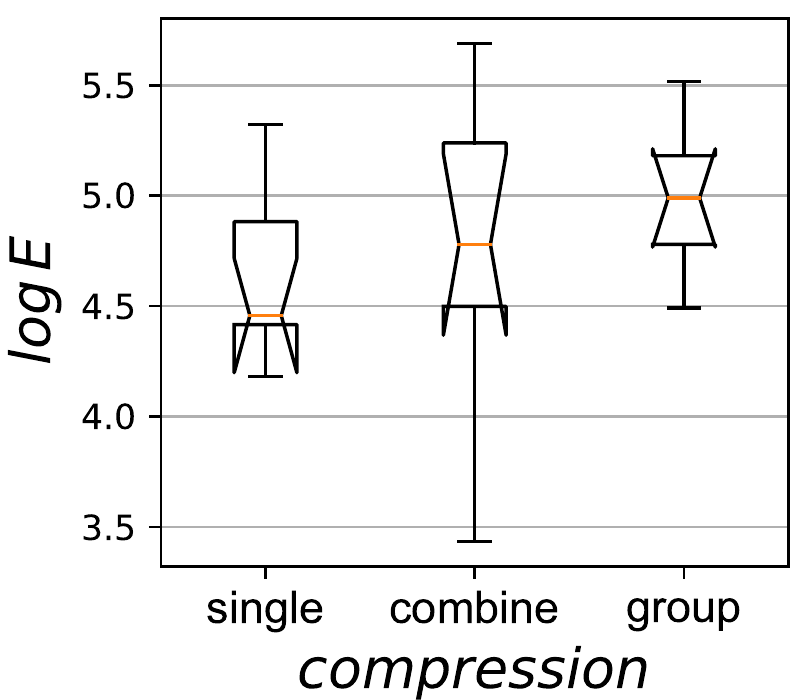}}
  \hfill
  \subfloat[]{
  \includegraphics[height=0.21\textwidth]{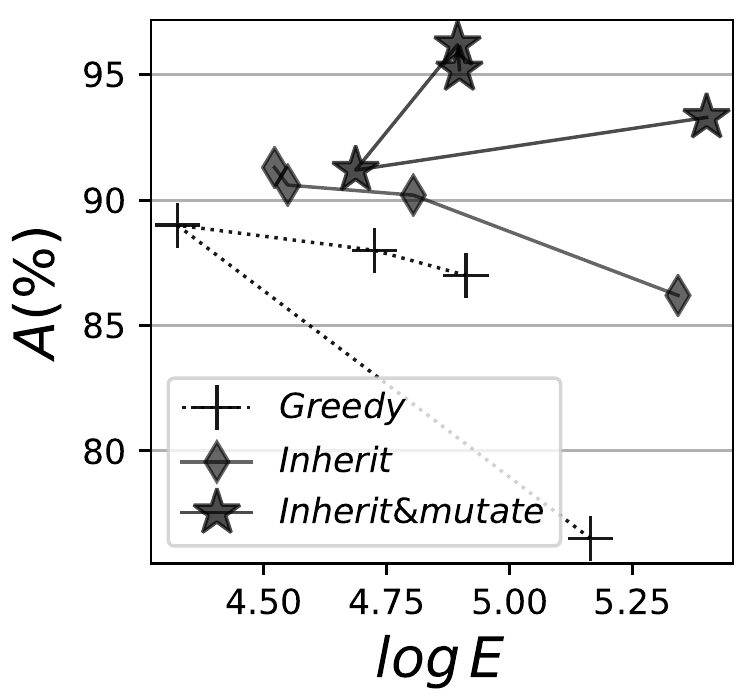}}
    \hfill
  \subfloat[]{
  \includegraphics[height=0.21\textwidth]{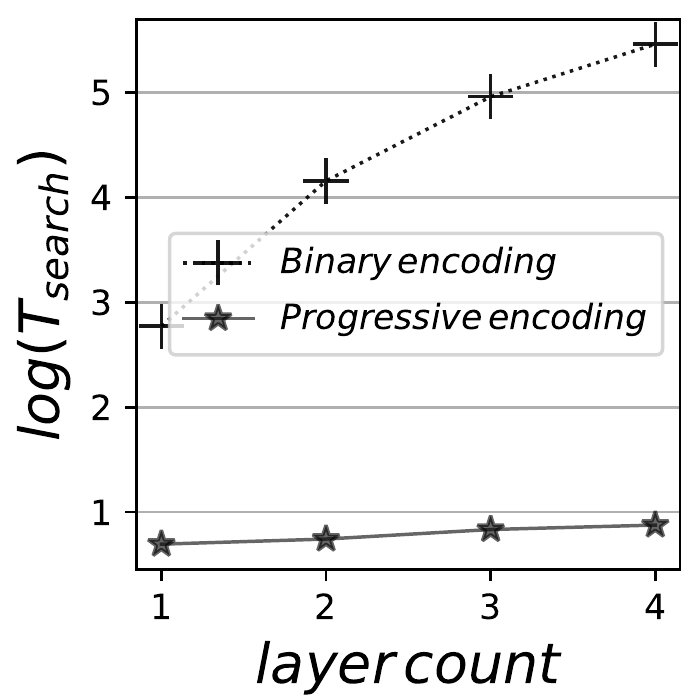}}
    \hfill
  \subfloat[]{
  \includegraphics[height=0.21\textwidth]{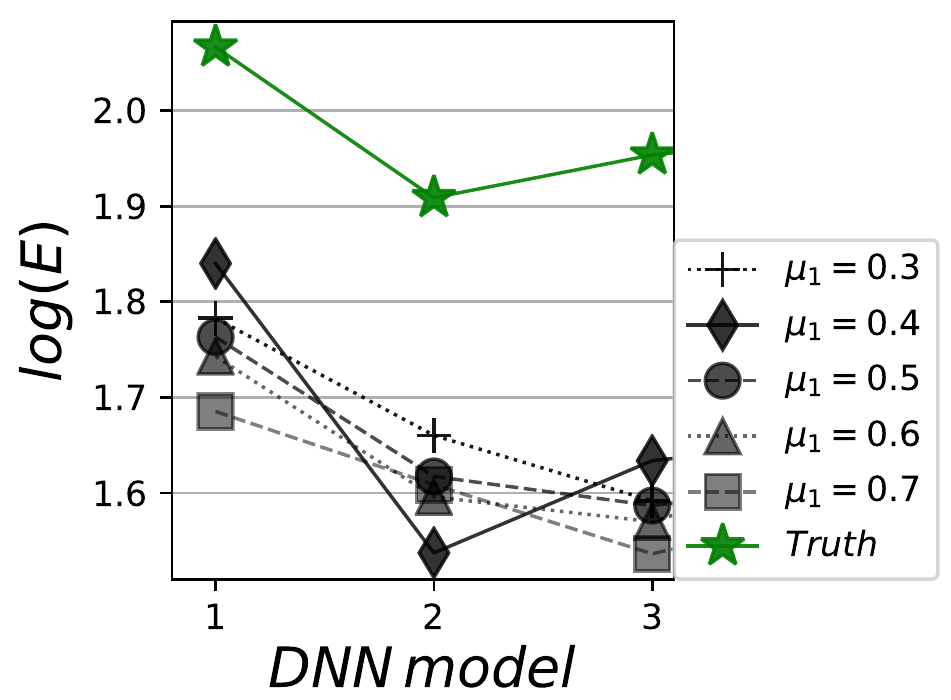}}
  \vspace{-1mm}
\caption{Impact of (a) hardware efficiency-guided combination, (b) layer-dependent inherit and mutation, (c) progressive shortest encoding, and (d) aggregation coefficients in arithmetic intensity on \systemnameposs performance.
}
\label{fig_benchmark}
\vspace{-0mm}
\end{figure}

\textbf{Performance}. Figure \ref{fig_exp_devices} summarizes the performance of DNN specialized by \systemname along with the dynamic changes of deployment context.
We first initialize the different DNN configurations for various platform constraints and then leverage \systemname to update the compression operator-variant selections according to the specific platform's dynamic contexts. 
We select four points of dynamic contexts.
\systemname identifies DNN configurations to obtain latency of $1.1 \sim 1.8$, parameter arithmetic intensity $81 \sim 151 $, and activation arithmetic intensity $192 \sim 397$ while have slightly degraded or even better accuracy $91.2\% \sim 98.6\%$.
We pick a time fragment to show four moments with dynamic deployment contexts, as shown in Table \ref{tb_exp_device_context}. 
As the gradual reduction of battery power and the dynamic fluctuation of Cache capacity, we show the performance changes of the DNN, which is continually scaled-down/up by selecting and combining different compression operators.
Moreover, \systemname supports scale up the model architecture again when the dynamic constraints on resource efficiency are relaxed, bringing better flexibility.

\textbf{Summary}. \systemname adaptively selects the proper combination of compression operators to optimize DNN performance continually to meet dynamic context demands. Moreover, \systemname accomplishes a flexible evolution, \ie support both scaling up and down the DNN configurations as the context demands change.

\subsection{Micro-benchmarks of \systemname}
\label{subsec_exp_bench}
In this subsection, we evaluate the impact of different factors on \systemnameposs design.

\subsubsection{Hardware Efficiency-guided Combination}
\label{subsubsec:exp_group}
We compare the performance of DNNs reconfigured by a stand-alone compression technique (\eg the Fire module~\cite{bib:iandola2016squeezenet}),  the blindly combined two compression techniques (\eg Fire module plus depth-wise pruning), and the proposed hardware-efficient grouping of compression operators (see Figure \ref{fig_benchmark}(a)).
And we show that the hardware-efficient grouping can always guarantee a comparable overall performance in terms of accuracy, energy efficiency and latency.

\begin{figure}[t]
  \centering
  \subfloat[Sound appears]{
  \includegraphics[height=0.15\textwidth]{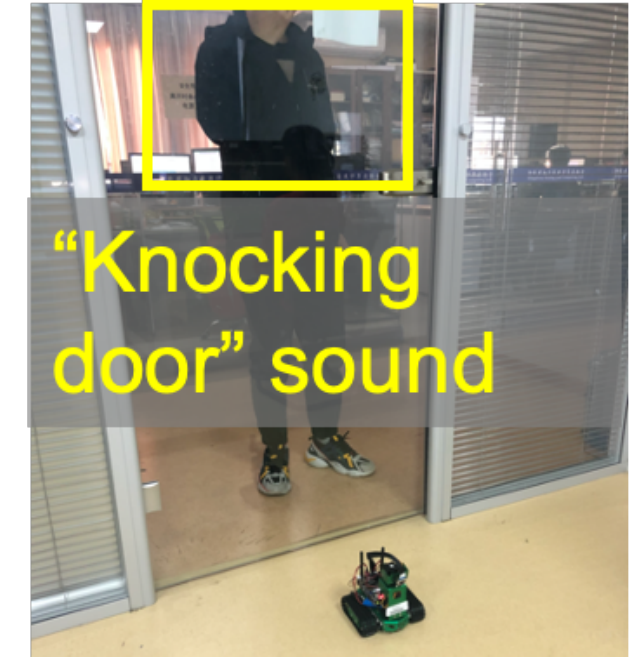}}
  \hfill
  \subfloat[DNN infers]{
  \includegraphics[height=0.15\textwidth]{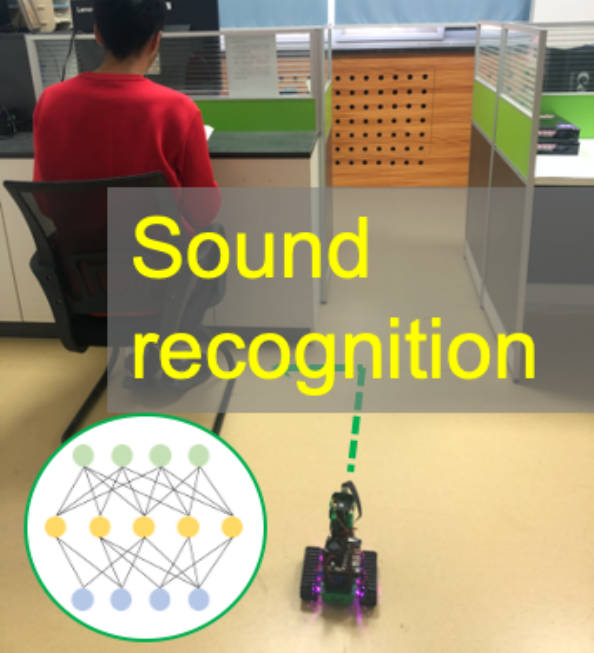}}
  \hfill
  \subfloat[Jetbot notifies]{
  \includegraphics[height=0.15\textwidth]{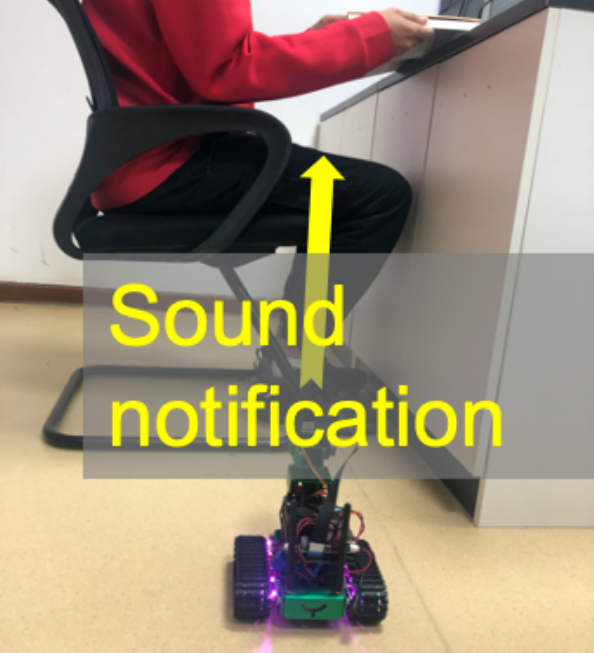}}
  \hfill
  \subfloat[NVIDIA Jetbot platform]{
  \includegraphics[height=0.15\textwidth]{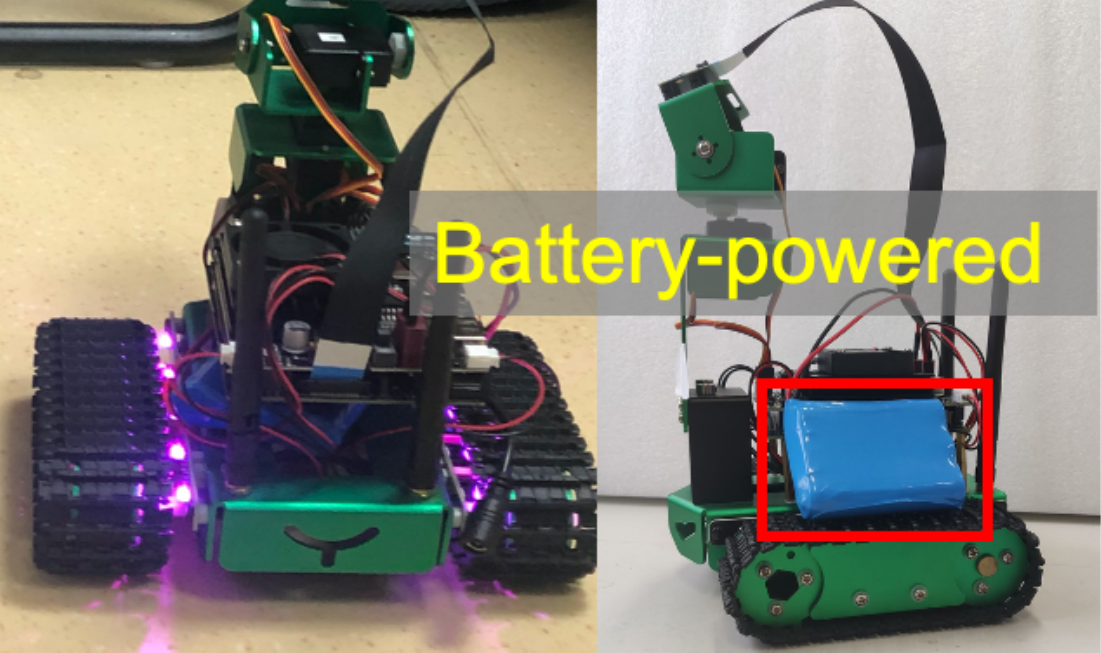}}
\caption{Case study of the DNN-based sound recognition application continuously, which runs from $9:00 am \sim 17:00 pm$ to provide sound awareness and notification service for the hard-of-hearing user.
}
\label{fig_case}
\vspace{-1mm}
\end{figure}

\begin{figure}[t]
  \centering
  \subfloat[Cache occupy]{
  \includegraphics[height=0.22\textwidth]{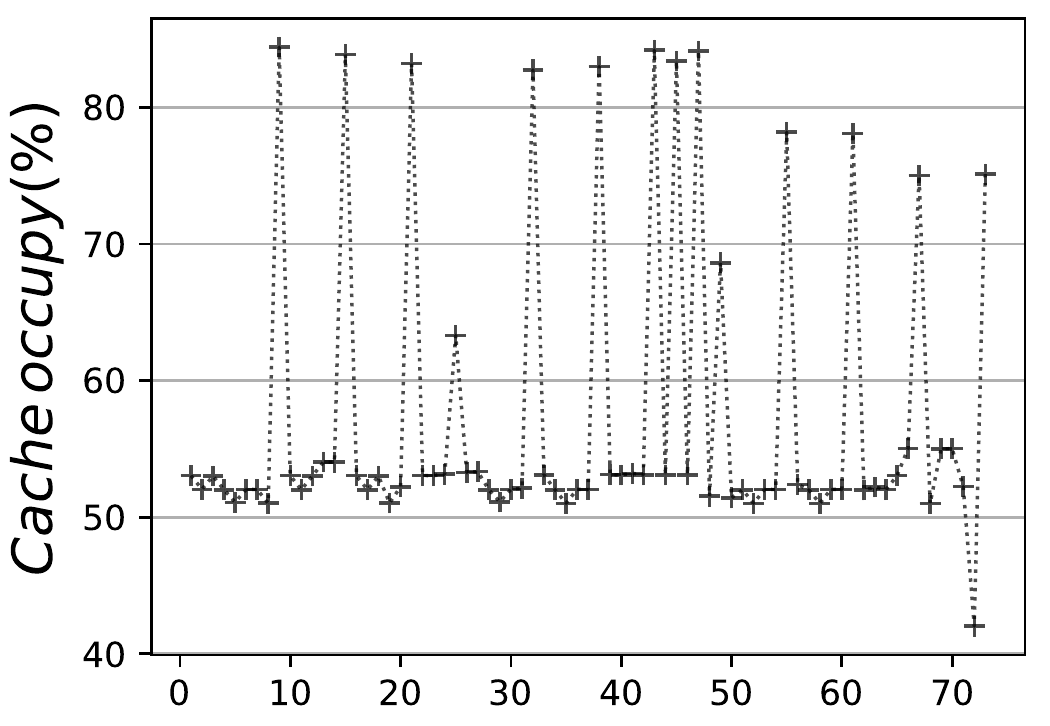}}
  \hfill
  \subfloat[Latency]{
  \includegraphics[height=0.22\textwidth]{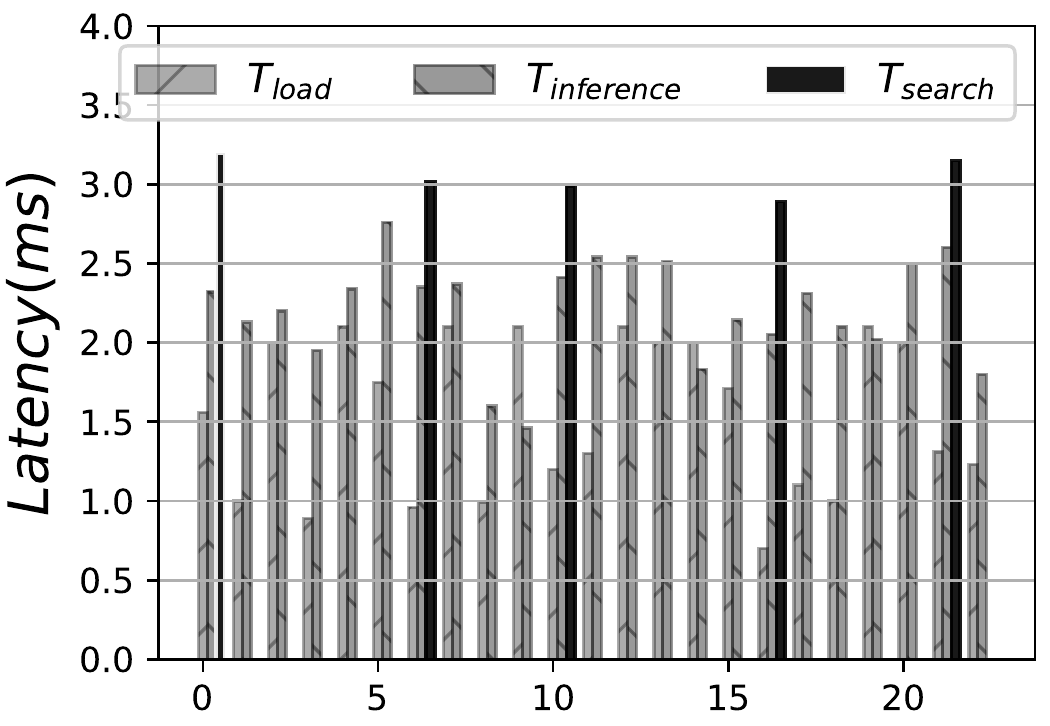}}
  \hfill
  \subfloat[DNN evolution performance]{
  \includegraphics[height=0.22\textwidth]{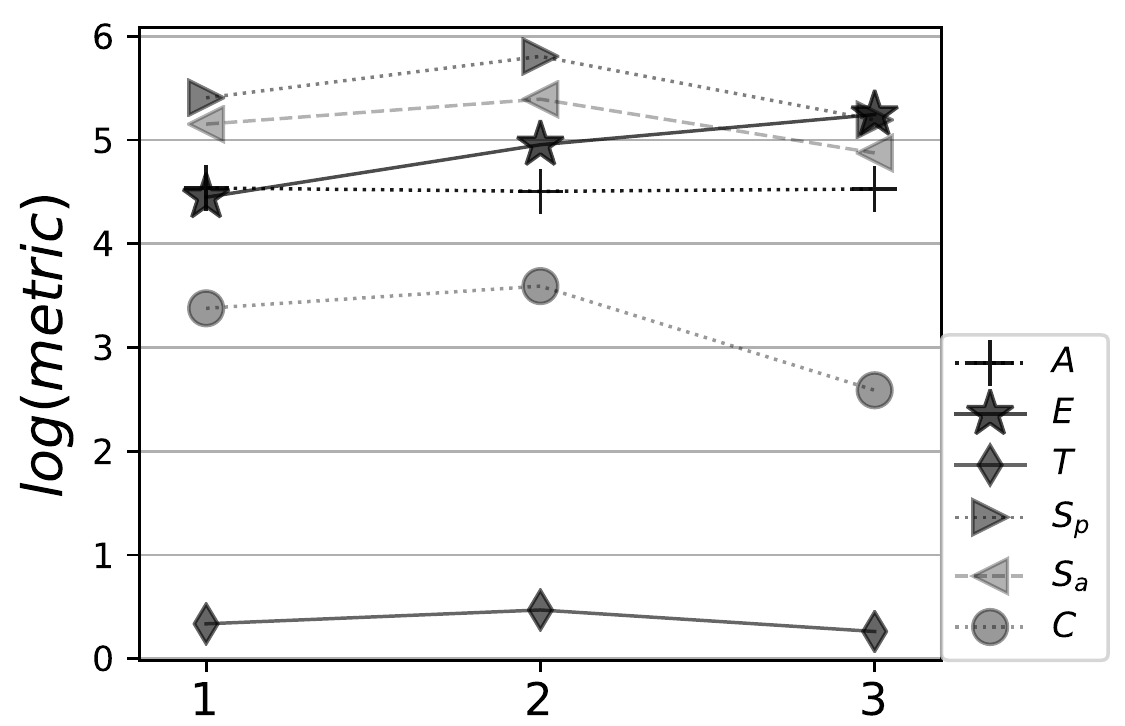}}
\caption{\systemname dynamically triggers the runtime adaptive DNN compression to satisfy the dynamic context.
}
\label{fig_case_exp}
\vspace{-1mm}
\end{figure}

\subsubsection{Layer-dependent Inheriting and Mutation}.
\label{subsubsec:exp_inherit}
As discussed in $\S$ \ref{subsec:search}, we leverage the inheriting and mutation schemes to balance the searching diversity and convergence.
We compare the locally greedy scheme layer by layer, the layer-dependent inheriting scheme, and the proposed layer-dependent inheriting plus mutation scheme in \systemname.
Figure \ref{fig_benchmark}(b) shows that \systemname achieves the best tradeoff between model accuracy and energy efficiency.

\subsubsection{Progressive Shortest Encoding}.
\label{subsubsec:exp_encoding}
The encoding of convolutional compression configurations at multiple layers affects the complexity of the search space.
Figure \ref{fig_benchmark}(c) compares the performance of classic binary encoding and progressive shortest encoding scheme.
And we find that \systemnameposs progressive shortest encoding method boosts the search efficiency.

\subsubsection{Aggregation Coefficients in Arithmetic Intensity}
\label{subsubsec:exp_coefficient}
As mentioned in $\S$ \ref{subsec:hardmetric}, the aggregation coefficients $\mu_1$ and $ \mu_2$ for the parameter and activation arithmetic intensity $\mu _1 C/S_p + \mu_2 C/S_a$ need to be optimized empirically.
Figure \ref{fig_benchmark}(d) illustrates the estimated energy consumption using different aggregation coefficient settings.
Therefore, we set $\mu_1=0.4, \mu_2=0.6$ by default across different platforms.

\subsection{Case Study}
\begin{figure*}[tb]
  \centering
  \includegraphics[width=.97\textwidth]{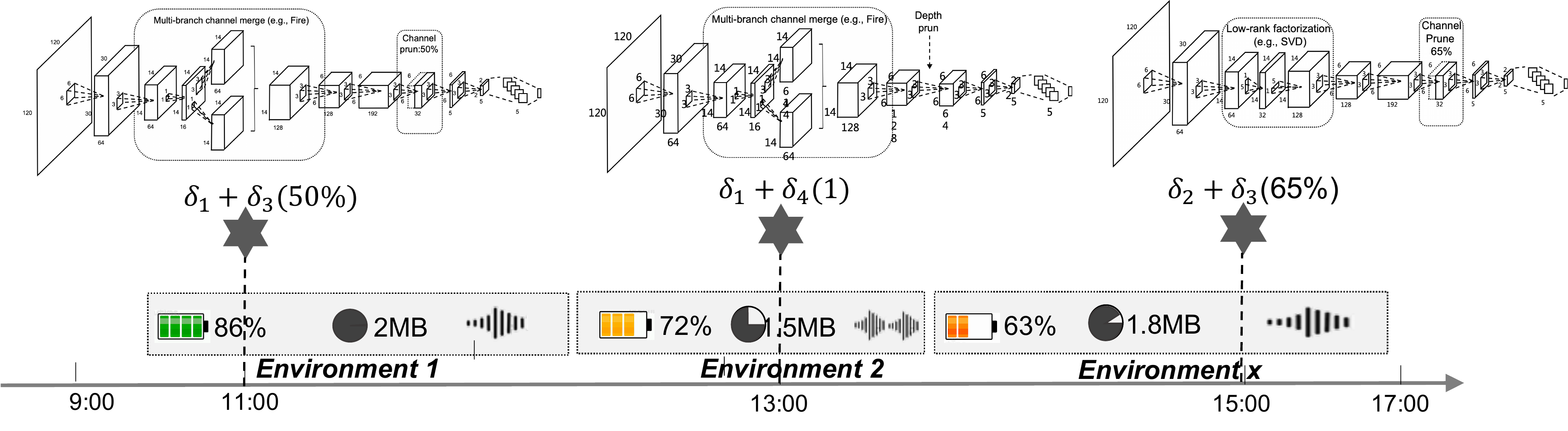}
  \caption{Context-adaptive and runtime-evolutionary DNN compression during dynamic inference.}
  \label{fig_context}
  \vspace{-4mm}
\end{figure*}

We deploy \systemname on a commercial mobile robot platform (\ie NVIDIA Jetbot, device4) and conduct a one-day experiment (09:00 to 17:00) to continually optimize the DNN configurations for a sound assistant application (\ie UbiEar~\cite{bib:sicong2017ubiear}).
This application adopts a DNN to realize a sound recognition and notification tool for hard-of-hearing people to sense emergency (\eg fire alarms, smoke alarms, kettle boiling whistle) and social events (\eg doorbell ring, knocking door, people crying). 
We simulate the dynamic mobile context of the DNN (as described in $\S$ \ref{sec:problem}) as follows. 
On the one hand, we artificially play some audio clips for emergency events and generate social events to control the happening frequency of acoustic events, affecting the DNN inference frequency.
%
On the other hand, we simulate the unpredictable storage resource contention by other software using the randomization noise (\eg Gaussian noise) $\sigma$ injection to the available capacity of L2-Cache, \ie $(2-\sigma)$MB.
Here, the maximum capacity of L2-Cache on NVIDIA Jetbot platform is $2MB$, and we update the randomized resource contention value of $\sigma$ per hour.
We do not artificially change the battery power, which is continuously consumed in the real-world as the application runs.
\rev{
Therefore, the remaining battery is dynamically changing, e.g., 86\%, 72\%, and 63\%, as shown in Figure \ref{fig_context}, which forms the dynamic energy budgets.
}

Figure \ref{fig_context} illustrates the dynamic deployment context (\ie energy, storage, event happening frequency) of the DNN for the continuous sound sensing application. 
The battery's remaining energy formulate the importance coefficient $\lambda_2$ in the runtime optimization problem (Equ. \ref{equ_opt}).
The available capacity of L2-Cache decides the storage budget of parameters $S_{bgt}$.
And the sound emergency frequency will indirectly influence the battery's power.
Different deployment contexts have various resource constraints and performance objective sensitivity, which lead to further performance and budget demands on the DNN.
\systemname triggers the runtime DNN evolution block by a pre-defined frequency (\eg every two hours) to shrink the DNN configurations in this regular day. 
Figure \ref{fig_case_exp}shows the runing performance of the DNNs specialized by \systemname.
%
\systemname can continually and adaptively select the best compression strategy to shrink the DNN configurations given diverse user demands.
\rev{
Specifically, it selects the $\delta_1$ (Fire) + $ \delta_3$ (pruning 50\% channel) for the regular resource-constrained moment, $\delta_1$ (Fire)+ $\delta_4$ (pruning 1 layer) for the tight memory constraint moment, and $\delta_2$ (SVD-based decomposition) + $\delta_3$ (pruning 65\% channel) for the tight battery-bounded moment.
}
The evolved models can achieve $\geq 95.6\%$ accuracy and $168.2 \sim 202.6$ arithmetic intensity.
\systemname searches the proper
combinations of compression operators that satisfy diverse demands on accuracy and resource efficiency within $2.8ms \sim 3.1ms$.

%% file: body/conclusion.tex
\section{Conclusion}
\label{sec:conclude}

This paper addressed the runtime adaptive DNN  compression problem to consider the \textit{dynamic deployment context} of continuously running mobile applications.
We present \systemname, a context-adaptive and runtime-evolutionary DNN compression framework that continually optimizes the DNN configurations (\ie architectures and weights) to adapt to the dynamic context.
We formulate the dynamic performance demands (\eg accuracy, latency, energy efficiency) as a time-varying constrained optimization problem.
And we propose a heuristic solution as quickly searching for the most suitable combination of retraining-free compression techniques at runtime.
To decouple DNN training from runtime adaptive compression, we put computation ahead in the training of a self-evolutionary network at design time (see $\S$ \ref{sec:offline}).
And we present the Runtime3C search algorithm and a set of searching speedup mechanisms to boost the runtime search efficiency and quality.
Evaluation using five different mobile applications across four mobile platforms and a real-world case study show the performance advantages of \systemname to evolve the DNN compression configurations locally online at millisecond level.
In the future work, facing the diverse and dynamic mobile scenarios (\eg data, task, and platform), more efforts and insights for the self-evolutionary deep model compression and optimization frameworks are much needed.
%

%


%% file: main.bbl

\begin{thebibliography}{64}


\ifx \showCODEN    \undefined \def \showCODEN     #1{\unskip}     \fi
\ifx \showDOI      \undefined \def \showDOI       #1{#1}\fi
\ifx \showISBNx    \undefined \def \showISBNx     #1{\unskip}     \fi
\ifx \showISBNxiii \undefined \def \showISBNxiii  #1{\unskip}     \fi
\ifx \showISSN     \undefined \def \showISSN      #1{\unskip}     \fi
\ifx \showLCCN     \undefined \def \showLCCN      #1{\unskip}     \fi
\ifx \shownote     \undefined \def \shownote      #1{#1}          \fi
\ifx \showarticletitle \undefined \def \showarticletitle #1{#1}   \fi
\ifx \showURL      \undefined \def \showURL       {\relax}        \fi
\providecommand\bibfield[2]{#2}
\providecommand\bibinfo[2]{#2}
\providecommand\natexlab[1]{#1}
\providecommand\showeprint[2][]{arXiv:#2}

\bibitem[\protect\citeauthoryear{Bender, Kindermans, Zoph, Vasudevan, and
  Le}{Bender et~al\mbox{.}}{2018}]%
        {bib:bender2018understanding}
\bibfield{author}{\bibinfo{person}{Gabriel Bender}, \bibinfo{person}{Pieter-Jan
  Kindermans}, \bibinfo{person}{Barret Zoph}, \bibinfo{person}{Vijay
  Vasudevan}, {and} \bibinfo{person}{Quoc Le}.}
  \bibinfo{year}{2018}\natexlab{}.
\newblock \showarticletitle{Understanding and simplifying one-shot architecture
  search}. In \bibinfo{booktitle}{\emph{International Conference on Machine
  Learning}}. \bibinfo{pages}{550--559}.
\newblock


\bibitem[\protect\citeauthoryear{Bhattacharya and Lane}{Bhattacharya and
  Lane}{2016}]%
        {bib:bhattacharya2016sparsification}
\bibfield{author}{\bibinfo{person}{Sourav Bhattacharya} {and}
  \bibinfo{person}{Nicholas~D Lane}.} \bibinfo{year}{2016}\natexlab{}.
\newblock \showarticletitle{Sparsification and separation of deep learning
  layers for constrained resource inference on wearables}. In
  \bibinfo{booktitle}{\emph{Proceedings of CD-ROM}}. \bibinfo{pages}{176--189}.
\newblock


\bibitem[\protect\citeauthoryear{Bhattacharya, Manousakas, Ramos, Venieris,
  Lane, and Mascolo}{Bhattacharya et~al\mbox{.}}{2020}]%
        {bib:bhattacharya2020countering}
\bibfield{author}{\bibinfo{person}{Sourav Bhattacharya},
  \bibinfo{person}{Dionysis Manousakas}, \bibinfo{person}{Alberto Gil~CP
  Ramos}, \bibinfo{person}{Stylianos~I Venieris}, \bibinfo{person}{Nicholas~D
  Lane}, {and} \bibinfo{person}{Cecilia Mascolo}.}
  \bibinfo{year}{2020}\natexlab{}.
\newblock \showarticletitle{Countering Acoustic Adversarial Attacks in
  Microphone-equipped Smart Home Devices}.
\newblock \bibinfo{journal}{\emph{Proceedings of the IMWUT}}
  \bibinfo{volume}{4}, \bibinfo{number}{2} (\bibinfo{year}{2020}),
  \bibinfo{pages}{1--24}.
\newblock


\bibitem[\protect\citeauthoryear{Cai, Chen, Zhang, Yu, and Wang}{Cai
  et~al\mbox{.}}{2017}]%
        {bib:cai2017efficient}
\bibfield{author}{\bibinfo{person}{Han Cai}, \bibinfo{person}{Tianyao Chen},
  \bibinfo{person}{Weinan Zhang}, \bibinfo{person}{Yong Yu}, {and}
  \bibinfo{person}{Jun Wang}.} \bibinfo{year}{2017}\natexlab{}.
\newblock \showarticletitle{Efficient architecture search by network
  transformation}.
\newblock \bibinfo{journal}{\emph{arXiv preprint arXiv:1707.04873}}
  (\bibinfo{year}{2017}).
\newblock


\bibitem[\protect\citeauthoryear{Cai, Gan, Wang, Zhang, and Han}{Cai
  et~al\mbox{.}}{2019}]%
        {bib:cai2019once}
\bibfield{author}{\bibinfo{person}{Han Cai}, \bibinfo{person}{Chuang Gan},
  \bibinfo{person}{Tianzhe Wang}, \bibinfo{person}{Zhekai Zhang}, {and}
  \bibinfo{person}{Song Han}.} \bibinfo{year}{2019}\natexlab{}.
\newblock \showarticletitle{Once-for-all: Train one network and specialize it
  for efficient deployment}.
\newblock \bibinfo{journal}{\emph{arXiv preprint arXiv:1908.09791}}
  (\bibinfo{year}{2019}).
\newblock


\bibitem[\protect\citeauthoryear{Cai, Zhu, and Han}{Cai et~al\mbox{.}}{2018}]%
        {bib:cai2018proxylessnas}
\bibfield{author}{\bibinfo{person}{Han Cai}, \bibinfo{person}{Ligeng Zhu},
  {and} \bibinfo{person}{Song Han}.} \bibinfo{year}{2018}\natexlab{}.
\newblock \showarticletitle{Proxylessnas: Direct neural architecture search on
  target task and hardware}.
\newblock \bibinfo{journal}{\emph{arXiv preprint arXiv:1812.00332}}
  (\bibinfo{year}{2018}).
\newblock


\bibitem[\protect\citeauthoryear{Chen, Choi, Yu, Han, and Chandraker}{Chen
  et~al\mbox{.}}{2017}]%
        {bib:chen2017learning}
\bibfield{author}{\bibinfo{person}{Guobin Chen}, \bibinfo{person}{Wongun Choi},
  \bibinfo{person}{Xiang Yu}, \bibinfo{person}{Tony Han}, {and}
  \bibinfo{person}{Manmohan Chandraker}.} \bibinfo{year}{2017}\natexlab{}.
\newblock \showarticletitle{Learning efficient object detection models with
  knowledge distillation}. In \bibinfo{booktitle}{\emph{Advances in Neural
  Information Processing Systems}}. \bibinfo{pages}{742--751}.
\newblock


\bibitem[\protect\citeauthoryear{Chen, Zhang, and Peng}{Chen
  et~al\mbox{.}}{2020}]%
        {bib:chen2020metier}
\bibfield{author}{\bibinfo{person}{Ling Chen}, \bibinfo{person}{Yi Zhang},
  {and} \bibinfo{person}{Liangying Peng}.} \bibinfo{year}{2020}\natexlab{}.
\newblock \showarticletitle{METIER: A Deep Multi-Task Learning Based Activity
  and User Recognition Model Using Wearable Sensors}.
\newblock \bibinfo{journal}{\emph{Proceedings of IMWUT}} \bibinfo{volume}{4},
  \bibinfo{number}{1} (\bibinfo{year}{2020}), \bibinfo{pages}{1--18}.
\newblock


\bibitem[\protect\citeauthoryear{Chen, Li, Bahsoon, and Yao}{Chen
  et~al\mbox{.}}{2018}]%
        {bib:chen2018femosaa}
\bibfield{author}{\bibinfo{person}{Tao Chen}, \bibinfo{person}{Ke Li},
  \bibinfo{person}{Rami Bahsoon}, {and} \bibinfo{person}{Xin Yao}.}
  \bibinfo{year}{2018}\natexlab{}.
\newblock \showarticletitle{FEMOSAA: Feature-guided and knee-driven
  multi-objective optimization for self-adaptive software}.
\newblock \bibinfo{journal}{\emph{ACM Transactions on Software Engineering and
  Methodology}} \bibinfo{volume}{27}, \bibinfo{number}{2}
  (\bibinfo{year}{2018}).
\newblock


\bibitem[\protect\citeauthoryear{Chen, Wilson, Tyree, Weinberger, and
  Chen}{Chen et~al\mbox{.}}{2016}]%
        {bib:chen2016compressing}
\bibfield{author}{\bibinfo{person}{Wenlin Chen}, \bibinfo{person}{James
  Wilson}, \bibinfo{person}{Stephen Tyree}, \bibinfo{person}{Kilian~Q
  Weinberger}, {and} \bibinfo{person}{Yixin Chen}.}
  \bibinfo{year}{2016}\natexlab{}.
\newblock \showarticletitle{Compressing convolutional neural networks in the
  frequency domain}. In \bibinfo{booktitle}{\emph{Proceedings of SIGKDD}}.
  \bibinfo{pages}{1475--1484}.
\newblock


\bibitem[\protect\citeauthoryear{Chen, Xie, Wu, and Tian}{Chen
  et~al\mbox{.}}{2019}]%
        {bib:chen2019progressive}
\bibfield{author}{\bibinfo{person}{Xin Chen}, \bibinfo{person}{Lingxi Xie},
  \bibinfo{person}{Jun Wu}, {and} \bibinfo{person}{Qi Tian}.}
  \bibinfo{year}{2019}\natexlab{}.
\newblock \showarticletitle{Progressive differentiable architecture search:
  Bridging the depth gap between search and evaluation}. In
  \bibinfo{booktitle}{\emph{Proceedings of ICCV}}. \bibinfo{pages}{1294--1303}.
\newblock


\bibitem[\protect\citeauthoryear{Cheng, Wang, Zhou, and Zhang}{Cheng
  et~al\mbox{.}}{2017}]%
        {bib:cheng2017survey}
\bibfield{author}{\bibinfo{person}{Yu Cheng}, \bibinfo{person}{Duo Wang},
  \bibinfo{person}{Pan Zhou}, {and} \bibinfo{person}{Tao Zhang}.}
  \bibinfo{year}{2017}\natexlab{}.
\newblock \showarticletitle{A survey of model compression and acceleration for
  deep neural networks}.
\newblock \bibinfo{journal}{\emph{arXiv preprint arXiv:1710.09282}}
  (\bibinfo{year}{2017}).
\newblock


\bibitem[\protect\citeauthoryear{Dai, Zhang, Wu, Yin, Sun, Wang, Dukhan, Hu,
  Wu, Jia, et~al\mbox{.}}{Dai et~al\mbox{.}}{2019}]%
        {bib:dai2019chamnet}
\bibfield{author}{\bibinfo{person}{Xiaoliang Dai}, \bibinfo{person}{Peizhao
  Zhang}, \bibinfo{person}{Bichen Wu}, \bibinfo{person}{Hongxu Yin},
  \bibinfo{person}{Fei Sun}, \bibinfo{person}{Yanghan Wang},
  \bibinfo{person}{Marat Dukhan}, \bibinfo{person}{Yunqing Hu},
  \bibinfo{person}{Yiming Wu}, \bibinfo{person}{Yangqing Jia}, {et~al\mbox{.}}}
  \bibinfo{year}{2019}\natexlab{}.
\newblock \showarticletitle{Chamnet: Towards efficient network design through
  platform-aware model adaptation}. In \bibinfo{booktitle}{\emph{Proceedings of
  CVPR}}. \bibinfo{pages}{11398--11407}.
\newblock


\bibitem[\protect\citeauthoryear{Deng, Dong, Socher, Li, Li, and Fei-Fei}{Deng
  et~al\mbox{.}}{2009}]%
        {data:imagenet}
\bibfield{author}{\bibinfo{person}{Jia Deng}, \bibinfo{person}{Wei Dong},
  \bibinfo{person}{Richard Socher}, \bibinfo{person}{Li-Jia Li},
  \bibinfo{person}{Kai Li}, {and} \bibinfo{person}{Li Fei-Fei}.}
  \bibinfo{year}{2009}\natexlab{}.
\newblock \showarticletitle{Imagenet: A large-scale hierarchical image
  database}.
\newblock In \bibinfo{booktitle}{\emph{Proceedings of CVPR}}.
\newblock


\bibitem[\protect\citeauthoryear{Fang, Sun, Peng, Zhang, Li, Liu, and
  Wang}{Fang et~al\mbox{.}}{2020}]%
        {bib:fang2020fast}
\bibfield{author}{\bibinfo{person}{Jiemin Fang}, \bibinfo{person}{Yuzhu Sun},
  \bibinfo{person}{Kangjian Peng}, \bibinfo{person}{Qian Zhang},
  \bibinfo{person}{Yuan Li}, \bibinfo{person}{Wenyu Liu}, {and}
  \bibinfo{person}{Xinggang Wang}.} \bibinfo{year}{2020}\natexlab{}.
\newblock \showarticletitle{Fast neural network adaptation via parameter
  remapping and architecture search}.
\newblock \bibinfo{journal}{\emph{arXiv preprint arXiv:2001.02525}}
  (\bibinfo{year}{2020}).
\newblock


\bibitem[\protect\citeauthoryear{French}{French}{1999}]%
        {bib:french1999catastrophic}
\bibfield{author}{\bibinfo{person}{Robert~M French}.}
  \bibinfo{year}{1999}\natexlab{}.
\newblock \showarticletitle{Catastrophic forgetting in connectionist networks}.
\newblock \bibinfo{journal}{\emph{Trends in cognitive sciences}}
  \bibinfo{volume}{3}, \bibinfo{number}{4} (\bibinfo{year}{1999}),
  \bibinfo{pages}{128--135}.
\newblock


\bibitem[\protect\citeauthoryear{Gao, Zhao, Dudziak, Mullins, and Xu}{Gao
  et~al\mbox{.}}{2018}]%
        {bib:gao2018dynamic}
\bibfield{author}{\bibinfo{person}{Xitong Gao}, \bibinfo{person}{Yiren Zhao},
  \bibinfo{person}{{\L}ukasz Dudziak}, \bibinfo{person}{Robert Mullins}, {and}
  \bibinfo{person}{Cheng-zhong Xu}.} \bibinfo{year}{2018}\natexlab{}.
\newblock \showarticletitle{Dynamic channel pruning: Feature boosting and
  suppression}.
\newblock \bibinfo{journal}{\emph{arXiv preprint arXiv:1810.05331}}
  (\bibinfo{year}{2018}).
\newblock


\bibitem[\protect\citeauthoryear{Gholami, Kwon, Wu, Tai, Yue, Jin, Zhao, and
  Keutzer}{Gholami et~al\mbox{.}}{2018}]%
        {bib:gholami2018squeezenext}
\bibfield{author}{\bibinfo{person}{Amir Gholami}, \bibinfo{person}{Kiseok
  Kwon}, \bibinfo{person}{Bichen Wu}, \bibinfo{person}{Zizheng Tai},
  \bibinfo{person}{Xiangyu Yue}, \bibinfo{person}{Peter Jin},
  \bibinfo{person}{Sicheng Zhao}, {and} \bibinfo{person}{Kurt Keutzer}.}
  \bibinfo{year}{2018}\natexlab{}.
\newblock \showarticletitle{Squeezenext: Hardware-aware neural network design}.
  In \bibinfo{booktitle}{\emph{Proceedings of CVPR}}.
  \bibinfo{pages}{1638--1647}.
\newblock


\bibitem[\protect\citeauthoryear{Google}{Google}{2017}]%
        {url:tensorflow}
\bibfield{author}{\bibinfo{person}{Google}.} \bibinfo{year}{2017}\natexlab{}.
\newblock \bibinfo{title}{TensorFlow}.
\newblock
\newblock
\newblock
\shownote{\url{https://goo.gl/j7HAZJ}.}


\bibitem[\protect\citeauthoryear{Han, Shen, Philipose, Agarwal, Wolman, and
  Krishnamurthy}{Han et~al\mbox{.}}{2016}]%
        {bib:han2016mcdnn}
\bibfield{author}{\bibinfo{person}{Seungyeop Han}, \bibinfo{person}{Haichen
  Shen}, \bibinfo{person}{Matthai Philipose}, \bibinfo{person}{Sharad Agarwal},
  \bibinfo{person}{Alec Wolman}, {and} \bibinfo{person}{Arvind Krishnamurthy}.}
  \bibinfo{year}{2016}\natexlab{}.
\newblock \showarticletitle{Mcdnn: An approximation-based execution framework
  for deep stream processing under resource constraints}. In
  \bibinfo{booktitle}{\emph{Proceedings of MobiSys}}.
  \bibinfo{pages}{123--136}.
\newblock


\bibitem[\protect\citeauthoryear{He, Zhang, Ren, and Sun}{He
  et~al\mbox{.}}{2016}]%
        {bib:he2016deep}
\bibfield{author}{\bibinfo{person}{Kaiming He}, \bibinfo{person}{Xiangyu
  Zhang}, \bibinfo{person}{Shaoqing Ren}, {and} \bibinfo{person}{Jian Sun}.}
  \bibinfo{year}{2016}\natexlab{}.
\newblock \showarticletitle{Deep residual learning for image recognition}. In
  \bibinfo{booktitle}{\emph{Proceedings of CVPR}}. \bibinfo{pages}{770--778}.
\newblock


\bibitem[\protect\citeauthoryear{He, Lin, Liu, Wang, Li, and Han}{He
  et~al\mbox{.}}{2018}]%
        {bib:he2018amc}
\bibfield{author}{\bibinfo{person}{Yihui He}, \bibinfo{person}{Ji Lin},
  \bibinfo{person}{Zhijian Liu}, \bibinfo{person}{Hanrui Wang},
  \bibinfo{person}{Li-Jia Li}, {and} \bibinfo{person}{Song Han}.}
  \bibinfo{year}{2018}\natexlab{}.
\newblock \showarticletitle{Amc: Automl for model compression and acceleration
  on mobile devices}. In \bibinfo{booktitle}{\emph{Proceedings of ECCV}}.
  \bibinfo{pages}{784--800}.
\newblock


\bibitem[\protect\citeauthoryear{He, Zhang, and Sun}{He et~al\mbox{.}}{2017}]%
        {bib:he2017channel}
\bibfield{author}{\bibinfo{person}{Yihui He}, \bibinfo{person}{Xiangyu Zhang},
  {and} \bibinfo{person}{Jian Sun}.} \bibinfo{year}{2017}\natexlab{}.
\newblock \showarticletitle{Channel pruning for accelerating very deep neural
  networks}. In \bibinfo{booktitle}{\emph{Proceedings of the IEEE International
  Conference on Computer Vision}}. \bibinfo{pages}{1389--1397}.
\newblock


\bibitem[\protect\citeauthoryear{Howard, Zhu, Chen, Kalenichenko, Wang, Weyand,
  Andreetto, and Adam}{Howard et~al\mbox{.}}{2017}]%
        {bib:howard2017mobilenets}
\bibfield{author}{\bibinfo{person}{Andrew~G Howard}, \bibinfo{person}{Menglong
  Zhu}, \bibinfo{person}{Bo Chen}, \bibinfo{person}{Dmitry Kalenichenko},
  \bibinfo{person}{Weijun Wang}, \bibinfo{person}{Tobias Weyand},
  \bibinfo{person}{Marco Andreetto}, {and} \bibinfo{person}{Hartwig Adam}.}
  \bibinfo{year}{2017}\natexlab{}.
\newblock \showarticletitle{Mobilenets: Efficient convolutional neural networks
  for mobile vision applications}.
\newblock \bibinfo{journal}{\emph{arXiv preprint arXiv:1704.04861}}
  (\bibinfo{year}{2017}).
\newblock


\bibitem[\protect\citeauthoryear{Iandola, Han, Moskewicz, Ashraf, Dally, and
  Keutzer}{Iandola et~al\mbox{.}}{2016}]%
        {bib:iandola2016squeezenet}
\bibfield{author}{\bibinfo{person}{Forrest~N Iandola}, \bibinfo{person}{Song
  Han}, \bibinfo{person}{Matthew~W Moskewicz}, \bibinfo{person}{Khalid Ashraf},
  \bibinfo{person}{William~J Dally}, {and} \bibinfo{person}{Kurt Keutzer}.}
  \bibinfo{year}{2016}\natexlab{}.
\newblock \showarticletitle{SqueezeNet: AlexNet-level accuracy with 50x fewer
  parameters and< 0.5 MB model size}.
\newblock \bibinfo{journal}{\emph{arXiv preprint arXiv:1602.07360}}
  (\bibinfo{year}{2016}).
\newblock


\bibitem[\protect\citeauthoryear{Jha and Mittal}{Jha and Mittal}{2020}]%
        {bib:jha2020modeling}
\bibfield{author}{\bibinfo{person}{Nandan~Kumar Jha} {and}
  \bibinfo{person}{Sparsh Mittal}.} \bibinfo{year}{2020}\natexlab{}.
\newblock \showarticletitle{Modeling Data Reuse in Deep Neural Networks by
  Taking Data-Types into Cognizance}.
\newblock \bibinfo{journal}{\emph{IEEE Trans. Comput.}} (\bibinfo{year}{2020}).
\newblock


\bibitem[\protect\citeauthoryear{Jha, Mittal, and Mattela}{Jha
  et~al\mbox{.}}{2019}]%
        {bib:jha2019ramifications}
\bibfield{author}{\bibinfo{person}{Nandan~Kumar Jha}, \bibinfo{person}{Sparsh
  Mittal}, {and} \bibinfo{person}{Govardhan Mattela}.}
  \bibinfo{year}{2019}\natexlab{}.
\newblock \showarticletitle{The ramifications of making deep neural networks
  compact}. In \bibinfo{booktitle}{\emph{Proceedings of VLSID}}. IEEE,
  \bibinfo{pages}{215--220}.
\newblock


\bibitem[\protect\citeauthoryear{Jiang, Hu, Xiao, Zhang, and Zhu}{Jiang
  et~al\mbox{.}}{2019}]%
        {bib:jiang2019improved}
\bibfield{author}{\bibinfo{person}{Yufan Jiang}, \bibinfo{person}{Chi Hu},
  \bibinfo{person}{Tong Xiao}, \bibinfo{person}{Chunliang Zhang}, {and}
  \bibinfo{person}{Jingbo Zhu}.} \bibinfo{year}{2019}\natexlab{}.
\newblock \showarticletitle{Improved differentiable architecture search for
  language modeling and named entity recognition}. In
  \bibinfo{booktitle}{\emph{Proceedings of EMNLP-IJCNLP}}.
  \bibinfo{pages}{3576--3581}.
\newblock


\bibitem[\protect\citeauthoryear{Kaggle}{Kaggle}{2019}]%
        {data:statefarm}
\bibfield{author}{\bibinfo{person}{Kaggle}.} \bibinfo{year}{2019}\natexlab{}.
\newblock \bibinfo{title}{State Farm Distracted Driver Detection}.
\newblock
  \bibinfo{howpublished}{\url{https://www.kaggle.com/c/state-farm-distracted-driver-detection}}.
\newblock


\bibitem[\protect\citeauthoryear{Kirkpatrick, Pascanu, Rabinowitz, Veness,
  Desjardins, Rusu, Milan, Quan, Ramalho, Grabska-Barwinska,
  et~al\mbox{.}}{Kirkpatrick et~al\mbox{.}}{2017}]%
        {bib:kirkpatrick2017overcoming}
\bibfield{author}{\bibinfo{person}{James Kirkpatrick}, \bibinfo{person}{Razvan
  Pascanu}, \bibinfo{person}{Neil Rabinowitz}, \bibinfo{person}{Joel Veness},
  \bibinfo{person}{Guillaume Desjardins}, \bibinfo{person}{Andrei~A Rusu},
  \bibinfo{person}{Kieran Milan}, \bibinfo{person}{John Quan},
  \bibinfo{person}{Tiago Ramalho}, \bibinfo{person}{Agnieszka
  Grabska-Barwinska}, {et~al\mbox{.}}} \bibinfo{year}{2017}\natexlab{}.
\newblock \showarticletitle{Overcoming catastrophic forgetting in neural
  networks}.
\newblock \bibinfo{journal}{\emph{Proceedings of PNAS}} \bibinfo{volume}{114},
  \bibinfo{number}{13} (\bibinfo{year}{2017}), \bibinfo{pages}{3521--3526}.
\newblock


\bibitem[\protect\citeauthoryear{Krizhevsky}{Krizhevsky}{2009a}]%
        {data:cifar}
\bibfield{author}{\bibinfo{person}{Alex Krizhevsky}.}
  \bibinfo{year}{2009}\natexlab{a}.
\newblock \bibinfo{title}{Learning multiple layers of features from tiny
  images}.
\newblock
  \bibinfo{howpublished}{\url{https://www.tensorflow.org/datasets/catalog/cifar100}}.
\newblock


\bibitem[\protect\citeauthoryear{Krizhevsky}{Krizhevsky}{2009b}]%
        {data:cifar100}
\bibfield{author}{\bibinfo{person}{Alex Krizhevsky}.}
  \bibinfo{year}{2009}\natexlab{b}.
\newblock \bibinfo{booktitle}{\emph{Learning multiple layers of features from
  tiny images}}.
\newblock \bibinfo{type}{{T}echnical {R}eport}.
\newblock


\bibitem[\protect\citeauthoryear{Krizhevsky, Sutskever, and Hinton}{Krizhevsky
  et~al\mbox{.}}{2012}]%
        {model:alexnet}
\bibfield{author}{\bibinfo{person}{Alex Krizhevsky}, \bibinfo{person}{Ilya
  Sutskever}, {and} \bibinfo{person}{Geoffrey~E Hinton}.}
  \bibinfo{year}{2012}\natexlab{}.
\newblock \showarticletitle{Imagenet classification with deep convolutional
  neural networks}. In \bibinfo{booktitle}{\emph{Advances in neural information
  processing systems}}. \bibinfo{pages}{1097--1105}.
\newblock


\bibitem[\protect\citeauthoryear{Kwon, Tong, Haresamudram, Gao, Abowd, Lane,
  and Ploetz}{Kwon et~al\mbox{.}}{2020}]%
        {bib:kwon2020imutube}
\bibfield{author}{\bibinfo{person}{Hyeokhyen Kwon}, \bibinfo{person}{Catherine
  Tong}, \bibinfo{person}{Harish Haresamudram}, \bibinfo{person}{Yan Gao},
  \bibinfo{person}{Gregory~D Abowd}, \bibinfo{person}{Nicholas~D Lane}, {and}
  \bibinfo{person}{Thomas Ploetz}.} \bibinfo{year}{2020}\natexlab{}.
\newblock \showarticletitle{IMUTube: Automatic extraction of virtual on-body
  accelerometry from video for human activity recognition}.
\newblock \bibinfo{journal}{\emph{arXiv preprint arXiv:2006.05675}}
  (\bibinfo{year}{2020}).
\newblock


\bibitem[\protect\citeauthoryear{Lane, Bhattacharya, Georgiev, Forlivesi, Jiao,
  Qendro, and Kawsar}{Lane et~al\mbox{.}}{2016}]%
        {bib:lane2016deepx}
\bibfield{author}{\bibinfo{person}{Nicholas~D Lane}, \bibinfo{person}{Sourav
  Bhattacharya}, \bibinfo{person}{Petko Georgiev}, \bibinfo{person}{Claudio
  Forlivesi}, \bibinfo{person}{Lei Jiao}, \bibinfo{person}{Lorena Qendro},
  {and} \bibinfo{person}{Fahim Kawsar}.} \bibinfo{year}{2016}\natexlab{}.
\newblock \showarticletitle{Deepx: A software accelerator for low-power deep
  learning inference on mobile devices}. In
  \bibinfo{booktitle}{\emph{Proceedings of IPSN}}. IEEE,
  \bibinfo{pages}{1--12}.
\newblock


\bibitem[\protect\citeauthoryear{Lane, Georgiev, and Qendro}{Lane
  et~al\mbox{.}}{2015}]%
        {bib:lane2015deepear}
\bibfield{author}{\bibinfo{person}{Nicholas~D Lane}, \bibinfo{person}{Petko
  Georgiev}, {and} \bibinfo{person}{Lorena Qendro}.}
  \bibinfo{year}{2015}\natexlab{}.
\newblock \showarticletitle{DeepEar: robust smartphone audio sensing in
  unconstrained acoustic environments using deep learning}. In
  \bibinfo{booktitle}{\emph{Proceedings of the 2015 ACM International Joint
  Conference on Pervasive and Ubiquitous Computing}}.
  \bibinfo{pages}{283--294}.
\newblock


\bibitem[\protect\citeauthoryear{Li, Yun, Kim, and Kim}{Li
  et~al\mbox{.}}{2019}]%
        {bib:li2019dabnet}
\bibfield{author}{\bibinfo{person}{Gen Li}, \bibinfo{person}{Inyoung Yun},
  \bibinfo{person}{Jonghyun Kim}, {and} \bibinfo{person}{Joongkyu Kim}.}
  \bibinfo{year}{2019}\natexlab{}.
\newblock \showarticletitle{Dabnet: Depth-wise asymmetric bottleneck for
  real-time semantic segmentation}.
\newblock \bibinfo{journal}{\emph{arXiv preprint arXiv:1907.11357}}
  (\bibinfo{year}{2019}).
\newblock


\bibitem[\protect\citeauthoryear{Li, Zhang, Chen, and Smola}{Li
  et~al\mbox{.}}{2014}]%
        {bib:li2014efficient}
\bibfield{author}{\bibinfo{person}{Mu Li}, \bibinfo{person}{Tong Zhang},
  \bibinfo{person}{Yuqiang Chen}, {and} \bibinfo{person}{Alexander~J Smola}.}
  \bibinfo{year}{2014}\natexlab{}.
\newblock \showarticletitle{Efficient mini-batch training for stochastic
  optimization}. In \bibinfo{booktitle}{\emph{Proceedings of SIGKDD}}.
  \bibinfo{pages}{661--670}.
\newblock


\bibitem[\protect\citeauthoryear{Liu, Simonyan, Vinyals, Fernando, and
  Kavukcuoglu}{Liu et~al\mbox{.}}{2017}]%
        {bib:liu2017hierarchical}
\bibfield{author}{\bibinfo{person}{Hanxiao Liu}, \bibinfo{person}{Karen
  Simonyan}, \bibinfo{person}{Oriol Vinyals}, \bibinfo{person}{Chrisantha
  Fernando}, {and} \bibinfo{person}{Koray Kavukcuoglu}.}
  \bibinfo{year}{2017}\natexlab{}.
\newblock \showarticletitle{Hierarchical representations for efficient
  architecture search}.
\newblock \bibinfo{journal}{\emph{arXiv preprint arXiv:1711.00436}}
  (\bibinfo{year}{2017}).
\newblock


\bibitem[\protect\citeauthoryear{Liu, Simonyan, and Yang}{Liu
  et~al\mbox{.}}{2018b}]%
        {bib:liu2018darts}
\bibfield{author}{\bibinfo{person}{Hanxiao Liu}, \bibinfo{person}{Karen
  Simonyan}, {and} \bibinfo{person}{Yiming Yang}.}
  \bibinfo{year}{2018}\natexlab{b}.
\newblock \showarticletitle{Darts: Differentiable architecture search}.
\newblock \bibinfo{journal}{\emph{arXiv preprint arXiv:1806.09055}}
  (\bibinfo{year}{2018}).
\newblock


\bibitem[\protect\citeauthoryear{Liu, Du, Nan, Wang, Lin, et~al\mbox{.}}{Liu
  et~al\mbox{.}}{2020}]%
        {bib:liu2020adadeep}
\bibfield{author}{\bibinfo{person}{Sicong Liu}, \bibinfo{person}{Junzhao Du},
  \bibinfo{person}{Kaiming Nan}, \bibinfo{person}{Atlas Wang},
  \bibinfo{person}{Yingyan Lin}, {et~al\mbox{.}}}
  \bibinfo{year}{2020}\natexlab{}.
\newblock \showarticletitle{AdaDeep: A Usage-Driven, Automated Deep Model
  Compression Framework for Enabling Ubiquitous Intelligent Mobiles}.
\newblock \bibinfo{journal}{\emph{arXiv preprint arXiv:2006.04432}}
  (\bibinfo{year}{2020}).
\newblock


\bibitem[\protect\citeauthoryear{Liu, Lin, Zhou, Nan, Liu, and Du}{Liu
  et~al\mbox{.}}{2018a}]%
        {bib:liu2018demand}
\bibfield{author}{\bibinfo{person}{Sicong Liu}, \bibinfo{person}{Yingyan Lin},
  \bibinfo{person}{Zimu Zhou}, \bibinfo{person}{Kaiming Nan},
  \bibinfo{person}{Hui Liu}, {and} \bibinfo{person}{Junzhao Du}.}
  \bibinfo{year}{2018}\natexlab{a}.
\newblock \showarticletitle{On-demand deep model compression for mobile
  devices: A usage-driven model selection framework}. In
  \bibinfo{booktitle}{\emph{Proceedings of MobiSys}}.
  \bibinfo{pages}{389--400}.
\newblock


\bibitem[\protect\citeauthoryear{Luo and Wu}{Luo and Wu}{2020}]%
        {bib:luo2020autopruner}
\bibfield{author}{\bibinfo{person}{Jian-Hao Luo} {and} \bibinfo{person}{Jianxin
  Wu}.} \bibinfo{year}{2020}\natexlab{}.
\newblock \showarticletitle{Autopruner: An end-to-end trainable filter pruning
  method for efficient deep model inference}.
\newblock \bibinfo{journal}{\emph{Pattern Recognition}} (\bibinfo{year}{2020}),
  \bibinfo{pages}{107461}.
\newblock


\bibitem[\protect\citeauthoryear{Ren, Xiao, Chang, Huang, Li, Chen, and
  Wang}{Ren et~al\mbox{.}}{2020}]%
        {bib:ren2020comprehensive}
\bibfield{author}{\bibinfo{person}{Pengzhen Ren}, \bibinfo{person}{Yun Xiao},
  \bibinfo{person}{Xiaojun Chang}, \bibinfo{person}{Po-Yao Huang},
  \bibinfo{person}{Zhihui Li}, \bibinfo{person}{Xiaojiang Chen}, {and}
  \bibinfo{person}{Xin Wang}.} \bibinfo{year}{2020}\natexlab{}.
\newblock \showarticletitle{A Comprehensive Survey of Neural Architecture
  Search: Challenges and Solutions}.
\newblock \bibinfo{journal}{\emph{arXiv preprint arXiv:2006.02903}}
  (\bibinfo{year}{2020}).
\newblock


\bibitem[\protect\citeauthoryear{Saikia, Marrakchi, Zela, Hutter, and
  Brox}{Saikia et~al\mbox{.}}{2019}]%
        {bib:saikia2019autodispnet}
\bibfield{author}{\bibinfo{person}{Tonmoy Saikia}, \bibinfo{person}{Yassine
  Marrakchi}, \bibinfo{person}{Arber Zela}, \bibinfo{person}{Frank Hutter},
  {and} \bibinfo{person}{Thomas Brox}.} \bibinfo{year}{2019}\natexlab{}.
\newblock \showarticletitle{Autodispnet: Improving disparity estimation with
  automl}. In \bibinfo{booktitle}{\emph{Proceedings of ICCV}}.
  \bibinfo{pages}{1812--1823}.
\newblock


\bibitem[\protect\citeauthoryear{Sandler, Howard, Zhu, Zhmoginov, and
  Chen}{Sandler et~al\mbox{.}}{2018}]%
        {bib:sandler2018mobilenetv2}
\bibfield{author}{\bibinfo{person}{Mark Sandler}, \bibinfo{person}{Andrew
  Howard}, \bibinfo{person}{Menglong Zhu}, \bibinfo{person}{Andrey Zhmoginov},
  {and} \bibinfo{person}{Liang-Chieh Chen}.} \bibinfo{year}{2018}\natexlab{}.
\newblock \showarticletitle{Mobilenetv2: Inverted residuals and linear
  bottlenecks}. In \bibinfo{booktitle}{\emph{Proceedings of the IEEE conference
  on computer vision and pattern recognition}}. \bibinfo{pages}{4510--4520}.
\newblock


\bibitem[\protect\citeauthoryear{Sicong, Zimu, Junzhao, Longfei, Han, and
  Wang}{Sicong et~al\mbox{.}}{2017}]%
        {bib:sicong2017ubiear}
\bibfield{author}{\bibinfo{person}{Liu Sicong}, \bibinfo{person}{Zhou Zimu},
  \bibinfo{person}{Du Junzhao}, \bibinfo{person}{Shangguan Longfei},
  \bibinfo{person}{Jun Han}, {and} \bibinfo{person}{Xin Wang}.}
  \bibinfo{year}{2017}\natexlab{}.
\newblock \showarticletitle{Ubiear: Bringing location-independent sound
  awareness to the hard-of-hearing people with smartphones}.
\newblock \bibinfo{journal}{\emph{Proceedings of IMWUT}} \bibinfo{volume}{1},
  \bibinfo{number}{2} (\bibinfo{year}{2017}), \bibinfo{pages}{1--21}.
\newblock


\bibitem[\protect\citeauthoryear{Singh, Verma, Rai, and Namboodiri}{Singh
  et~al\mbox{.}}{2019}]%
        {bib:singh2019play}
\bibfield{author}{\bibinfo{person}{Pravendra Singh},
  \bibinfo{person}{Vinay~Kumar Verma}, \bibinfo{person}{Piyush Rai}, {and}
  \bibinfo{person}{Vinay~P Namboodiri}.} \bibinfo{year}{2019}\natexlab{}.
\newblock \showarticletitle{Play and prune: Adaptive filter pruning for deep
  model compression}.
\newblock \bibinfo{journal}{\emph{arXiv preprint arXiv:1905.04446}}
  (\bibinfo{year}{2019}).
\newblock


\bibitem[\protect\citeauthoryear{Tan, Chen, Pang, Vasudevan, Sandler, Howard,
  and Le}{Tan et~al\mbox{.}}{2019}]%
        {bib:tan2019mnasnet}
\bibfield{author}{\bibinfo{person}{Mingxing Tan}, \bibinfo{person}{Bo Chen},
  \bibinfo{person}{Ruoming Pang}, \bibinfo{person}{Vijay Vasudevan},
  \bibinfo{person}{Mark Sandler}, \bibinfo{person}{Andrew Howard}, {and}
  \bibinfo{person}{Quoc~V Le}.} \bibinfo{year}{2019}\natexlab{}.
\newblock \showarticletitle{Mnasnet: Platform-aware neural architecture search
  for mobile}. In \bibinfo{booktitle}{\emph{Proceedings of CVPR}}.
  \bibinfo{pages}{2820--2828}.
\newblock


\bibitem[\protect\citeauthoryear{Teerapittayanon, McDanel, and
  Kung}{Teerapittayanon et~al\mbox{.}}{2016}]%
        {bib:teerapittayanon2016branchynet}
\bibfield{author}{\bibinfo{person}{Surat Teerapittayanon},
  \bibinfo{person}{Bradley McDanel}, {and} \bibinfo{person}{Hsiang-Tsung
  Kung}.} \bibinfo{year}{2016}\natexlab{}.
\newblock \showarticletitle{Branchynet: Fast inference via early exiting from
  deep neural networks}. In \bibinfo{booktitle}{\emph{Proceedings of ICPR}}.
  IEEE, \bibinfo{pages}{2464--2469}.
\newblock


\bibitem[\protect\citeauthoryear{UCI}{UCI}{2017}]%
        {data:Har}
\bibfield{author}{\bibinfo{person}{UCI}.} \bibinfo{year}{2017}\natexlab{}.
\newblock \bibinfo{title}{Dataset for Human Activity Recognition}.
\newblock \bibinfo{howpublished}{\url{https://goo.gl/m5bRo1}}.
\newblock


\bibitem[\protect\citeauthoryear{Wang, Han, Leung, Niyato, Yan, and Chen}{Wang
  et~al\mbox{.}}{2020}]%
        {bib:wang2020convergence}
\bibfield{author}{\bibinfo{person}{Xiaofei Wang}, \bibinfo{person}{Yiwen Han},
  \bibinfo{person}{Victor~CM Leung}, \bibinfo{person}{Dusit Niyato},
  \bibinfo{person}{Xueqiang Yan}, {and} \bibinfo{person}{Xu Chen}.}
  \bibinfo{year}{2020}\natexlab{}.
\newblock \showarticletitle{Convergence of edge computing and deep learning: A
  comprehensive survey}.
\newblock \bibinfo{journal}{\emph{IEEE Communications Surveys \& Tutorials}}
  \bibinfo{volume}{22}, \bibinfo{number}{2} (\bibinfo{year}{2020}),
  \bibinfo{pages}{869--904}.
\newblock


\bibitem[\protect\citeauthoryear{Wu, Wang, Wu, Wang, Veeraraghavan, and Lin}{Wu
  et~al\mbox{.}}{2018b}]%
        {bib:wu2018deep}
\bibfield{author}{\bibinfo{person}{Junru Wu}, \bibinfo{person}{Yue Wang},
  \bibinfo{person}{Zhenyu Wu}, \bibinfo{person}{Zhangyang Wang},
  \bibinfo{person}{Ashok Veeraraghavan}, {and} \bibinfo{person}{Yingyan Lin}.}
  \bibinfo{year}{2018}\natexlab{b}.
\newblock \showarticletitle{Deep $ k $-Means: Re-Training and Parameter Sharing
  with Harder Cluster Assignments for Compressing Deep Convolutions}.
\newblock \bibinfo{journal}{\emph{arXiv preprint arXiv:1806.09228}}
  (\bibinfo{year}{2018}).
\newblock


\bibitem[\protect\citeauthoryear{Wu, Nagarajan, Kumar, Rennie, Davis, Grauman,
  and Feris}{Wu et~al\mbox{.}}{2018a}]%
        {bib:wu2018blockdrop}
\bibfield{author}{\bibinfo{person}{Zuxuan Wu}, \bibinfo{person}{Tushar
  Nagarajan}, \bibinfo{person}{Abhishek Kumar}, \bibinfo{person}{Steven
  Rennie}, \bibinfo{person}{Larry~S Davis}, \bibinfo{person}{Kristen Grauman},
  {and} \bibinfo{person}{Rogerio Feris}.} \bibinfo{year}{2018}\natexlab{a}.
\newblock \showarticletitle{Blockdrop: Dynamic inference paths in residual
  networks}. In \bibinfo{booktitle}{\emph{Proceedings of CVPR}}.
  \bibinfo{pages}{8817--8826}.
\newblock


\bibitem[\protect\citeauthoryear{Yang, He, Cao, and Fan}{Yang
  et~al\mbox{.}}{2020}]%
        {bib:yang2020progressive}
\bibfield{author}{\bibinfo{person}{Li Yang}, \bibinfo{person}{Zhezhi He},
  \bibinfo{person}{Yu Cao}, {and} \bibinfo{person}{Deliang Fan}.}
  \bibinfo{year}{2020}\natexlab{}.
\newblock \showarticletitle{A Progressive Sub-Network Searching Framework for
  Dynamic Inference}.
\newblock \bibinfo{journal}{\emph{arXiv preprint arXiv:2009.05681}}
  (\bibinfo{year}{2020}).
\newblock


\bibitem[\protect\citeauthoryear{Yang, Chen, and Sze}{Yang
  et~al\mbox{.}}{2017}]%
        {bib:yang2017designing}
\bibfield{author}{\bibinfo{person}{Tien-Ju Yang}, \bibinfo{person}{Yu-Hsin
  Chen}, {and} \bibinfo{person}{Vivienne Sze}.}
  \bibinfo{year}{2017}\natexlab{}.
\newblock \showarticletitle{Designing energy-efficient convolutional neural
  networks using energy-aware pruning}. In
  \bibinfo{booktitle}{\emph{Proceedings of CVPR}}. \bibinfo{pages}{5687--5695}.
\newblock


\bibitem[\protect\citeauthoryear{Yang, Yu, Zheng, and Shi}{Yang
  et~al\mbox{.}}{2019}]%
        {bib:yang2019proxitalk}
\bibfield{author}{\bibinfo{person}{Zhican Yang}, \bibinfo{person}{Chun Yu},
  \bibinfo{person}{Fengshi Zheng}, {and} \bibinfo{person}{Yuanchun Shi}.}
  \bibinfo{year}{2019}\natexlab{}.
\newblock \showarticletitle{ProxiTalk: Activate Speech Input by Bringing
  Smartphone to the Mouth}.
\newblock \bibinfo{journal}{\emph{Proceedings of IMWUT}} \bibinfo{volume}{3},
  \bibinfo{number}{3} (\bibinfo{year}{2019}), \bibinfo{pages}{1--25}.
\newblock


\bibitem[\protect\citeauthoryear{Yao, Zhao, Zhang, Su, and Abdelzaher}{Yao
  et~al\mbox{.}}{2017}]%
        {bib:yao2017deepiot}
\bibfield{author}{\bibinfo{person}{Shuochao Yao}, \bibinfo{person}{Yiran Zhao},
  \bibinfo{person}{Aston Zhang}, \bibinfo{person}{Lu Su}, {and}
  \bibinfo{person}{Tarek Abdelzaher}.} \bibinfo{year}{2017}\natexlab{}.
\newblock \showarticletitle{Deepiot: Compressing deep neural network structures
  for sensing systems with a compressor-critic framework}. In
  \bibinfo{booktitle}{\emph{Proceedings of SenSys}}. \bibinfo{pages}{1--14}.
\newblock


\bibitem[\protect\citeauthoryear{Yu and Huang}{Yu and Huang}{2019}]%
        {bib:yu2019autoslim}
\bibfield{author}{\bibinfo{person}{Jiahui Yu} {and} \bibinfo{person}{Thomas
  Huang}.} \bibinfo{year}{2019}\natexlab{}.
\newblock \showarticletitle{AutoSlim: Towards One-Shot Architecture Search for
  Channel Numbers}.
\newblock \bibinfo{journal}{\emph{arXiv preprint arXiv:1903.11728}}
  (\bibinfo{year}{2019}).
\newblock


\bibitem[\protect\citeauthoryear{Zhao, Barijough, and Gerstlauer}{Zhao
  et~al\mbox{.}}{2018}]%
        {bib:zhao2018deepthings}
\bibfield{author}{\bibinfo{person}{Zhuoran Zhao},
  \bibinfo{person}{Kamyar~Mirzazad Barijough}, {and} \bibinfo{person}{Andreas
  Gerstlauer}.} \bibinfo{year}{2018}\natexlab{}.
\newblock \showarticletitle{DeepThings: Distributed adaptive deep learning
  inference on resource-constrained IoT edge clusters}.
\newblock \bibinfo{journal}{\emph{IEEE Transactions on Computer-Aided Design of
  Integrated Circuits and Systems}} \bibinfo{volume}{37}, \bibinfo{number}{11}
  (\bibinfo{year}{2018}), \bibinfo{pages}{2348--2359}.
\newblock


\bibitem[\protect\citeauthoryear{Zhong, Yan, Wu, Shao, and Liu}{Zhong
  et~al\mbox{.}}{2018}]%
        {bib:zhong2018practical}
\bibfield{author}{\bibinfo{person}{Zhao Zhong}, \bibinfo{person}{Junjie Yan},
  \bibinfo{person}{Wei Wu}, \bibinfo{person}{Jing Shao}, {and}
  \bibinfo{person}{Cheng-Lin Liu}.} \bibinfo{year}{2018}\natexlab{}.
\newblock \showarticletitle{Practical block-wise neural network architecture
  generation}. In \bibinfo{booktitle}{\emph{Proceedings of CVPR}}.
  \bibinfo{pages}{2423--2432}.
\newblock


\bibitem[\protect\citeauthoryear{Zhou, Xiong, Socher, and Hoi}{Zhou
  et~al\mbox{.}}{2020}]%
        {bib:zhou2020theory}
\bibfield{author}{\bibinfo{person}{Pan Zhou}, \bibinfo{person}{Caiming Xiong},
  \bibinfo{person}{Richard Socher}, {and} \bibinfo{person}{Steven~CH Hoi}.}
  \bibinfo{year}{2020}\natexlab{}.
\newblock \showarticletitle{Theory-inspired path-regularized differential
  network architecture search}.
\newblock \bibinfo{journal}{\emph{arXiv preprint arXiv:2006.16537}}
  (\bibinfo{year}{2020}).
\newblock


\bibitem[\protect\citeauthoryear{Zhu and Zabaras}{Zhu and Zabaras}{2018}]%
        {bib:zhu2018bayesian}
\bibfield{author}{\bibinfo{person}{Yinhao Zhu} {and} \bibinfo{person}{Nicholas
  Zabaras}.} \bibinfo{year}{2018}\natexlab{}.
\newblock \showarticletitle{Bayesian deep convolutional encoder--decoder
  networks for surrogate modeling and uncertainty quantification}.
\newblock \bibinfo{journal}{\emph{J. Comput. Phys.}}  \bibinfo{volume}{366}
  (\bibinfo{year}{2018}), \bibinfo{pages}{415--447}.
\newblock


\bibitem[\protect\citeauthoryear{Zoph, Vasudevan, Shlens, and Le}{Zoph
  et~al\mbox{.}}{2018}]%
        {bib:zoph2018learning}
\bibfield{author}{\bibinfo{person}{Barret Zoph}, \bibinfo{person}{Vijay
  Vasudevan}, \bibinfo{person}{Jonathon Shlens}, {and} \bibinfo{person}{Quoc~V
  Le}.} \bibinfo{year}{2018}\natexlab{}.
\newblock \showarticletitle{Learning transferable architectures for scalable
  image recognition}. In \bibinfo{booktitle}{\emph{Proceedings of CVPR}}.
  \bibinfo{pages}{8697--8710}.
\newblock


\end{thebibliography}
